\documentclass[acmtog, authorversion]{acmart} 
\graphicspath{{./images/}}
\usepackage[normalem]{ulem}
\usepackage{color,hyphenat,balance,booktabs}
\usepackage[fleqn,tbtags]{mathtools}
\usepackage{autonum}
\usepackage{overpic}
\usepackage{wrapfig}
\usepackage{contour}
\usepackage{enumerate}
\usepackage{physics}
\usepackage{amsmath}
\usepackage[ruled]{algorithm2e} 

\SetCommentSty{mycommfont}
\setcitestyle{square}
\DeclareMathOperator*{\argmin}{arg\,min}

\usepackage{amsthm}
\newtheorem*{theorem*}{Theorem}
\usepackage{subcaption}
\usepackage{makecell}

\setcopyright{acmcopyright}
\acmJournal{TOG}

\begin{document}
\title{Unsupervised Shape Completion via Deep Prior in the Neural Tangent Kernel Perspective}
\author{Lei Chu}
\authornote{Work partially done during an internship at Microsoft.} 
\affiliation{
  \institution{University of Hong Kong}
}
\affiliation{
	\institution{Microsoft Research Asia}
}
\author{Hao Pan}
\authornote{Corresponding author.\vspace{-1mm}}
\affiliation{
  \institution{Microsoft Research Asia}
}
\author{Wenping Wang}
\affiliation{
	\institution{Texas A\&M University}
}
\affiliation{
  \institution{University of Hong Kong}
}
\authorsaddresses{Lei Chu, University of Hong Kong, lchu@connect.hku.hk; Hao Pan, Microsoft Research Asia, haopan@microsoft.com; Wenping Wang, Texas A\&M University, wenping@cs.hku.hk. \vspace{-1mm} }

\begin{abstract}
We present a novel approach for completing and reconstructing 3D shapes from incomplete scanned data by using deep neural networks. 
Rather than being trained on supervised completion tasks and applied on a testing shape, the network is optimized from scratch on the single testing shape, to fully adapt to the shape and complete the missing data using contextual guidance from the known regions.
The ability to complete missing data by an untrained neural network is usually referred to as the \emph{deep prior}. In this paper, we interpret the deep prior from a neural tangent kernel (NTK) perspective and show that the completed shape patches by the trained CNN are naturally similar to existing patches, as they are proximate in the kernel feature space induced by NTK.
The interpretation allows us to design more efficient network structures and learning mechanisms for the shape completion and reconstruction task.
Being more aware of structural regularities than both traditional and other unsupervised learning-based reconstruction methods, our approach completes large missing regions with plausible shapes and complements supervised learning-based methods that use database priors by requiring no extra training data set and showing flexible adaptation to a particular shape instance.

\end{abstract}

%
%



%
%


\maketitle

\section{Introduction}
\label{sec:intro}

\begin{figure}
\centering
\includegraphics[width=\linewidth]{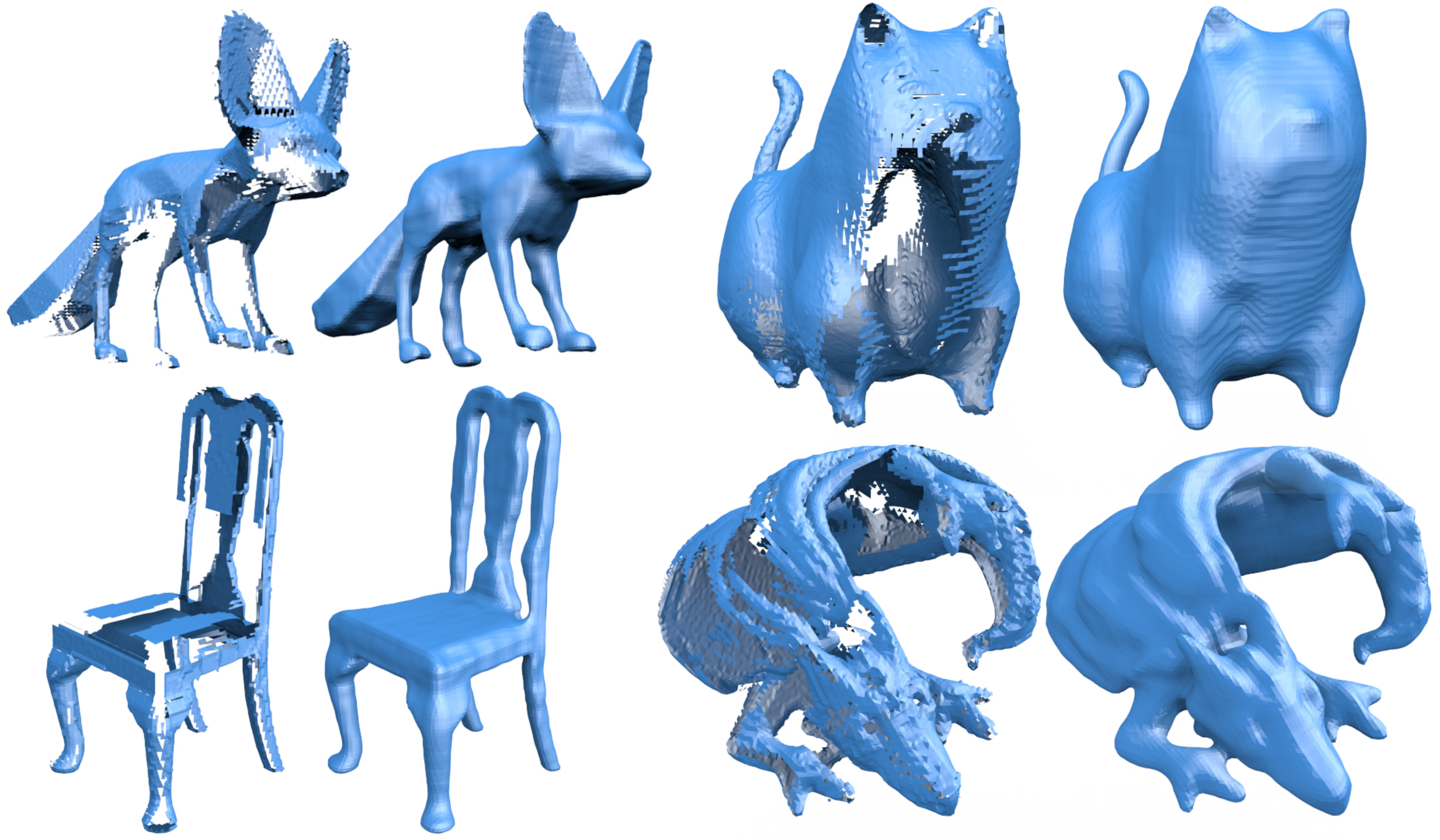}
\vspace{-7mm}
\caption{Completing partial scans of 3D shapes from different categories with large missing areas, by self-supervised training of sparse 3D CNNs that fit to the partial scans without using extra data set. Each example shows the partial scan (left) and the completed model (right).}
\label{fig:teaser}
\vspace{-5mm}
\end{figure}

Shape completion is an important task for reconstructing 3D shapes from scanned data, because frequently there are missing regions in the scanned data that are not well captured due to visibility or reflectance. 
Shape completion aims to provide a reasonable reconstruction of the missing regions, based on the context provided by the known regions.
The context can contain information in multiple levels of abstraction, e.g., starting from the object shape category, through its symmetry, down to the geometric texture and smoothness of the surface.
Correspondingly, numerous algorithms have been designed that use the various levels of information to guide completion, e.g., by fitting object class templates \cite{Huang2013,pauly2005example}, searching repetitive patterns and transferring to missing regions \cite{Bendels05,Harary:2014:CCS}, or minimizing shape variations for smoothness \cite{DavisDiffusion2002,kazhdan2006poisson}.
However, these hand-crafted rules are generally limited to the specific levels of geometric structures they are designed for but do not take advantage of the full context.

Deep learning based methods have been proposed to solve the completion task in a category-specific way \cite{wu20153d,3depn,HanComplete2017,Stutz2018CVPR,DeepSDF,dai2020sgnn}.
The benefit is that given enough training samples, strong structural prior can be learned for a particular category and guide reconstructing severely incomplete scans.
But it is not always possible to have a large amount of training samples which coherently have the desirable shape features, and a strong but limited prior may restrict the range of all possible shape variations recovered.
Recent works \cite{hanocka2020point2mesh,Williams_2019_CVPR} explore using structural priors of deep neural networks to reconstruct single 3D shapes without relying on data sets;
our approach follows a similar paradigm, but achieves better results for completion (Fig.~\ref{fig:comparison}).

\begin{figure*}
    \centering
    \begin{overpic}[width=\linewidth]{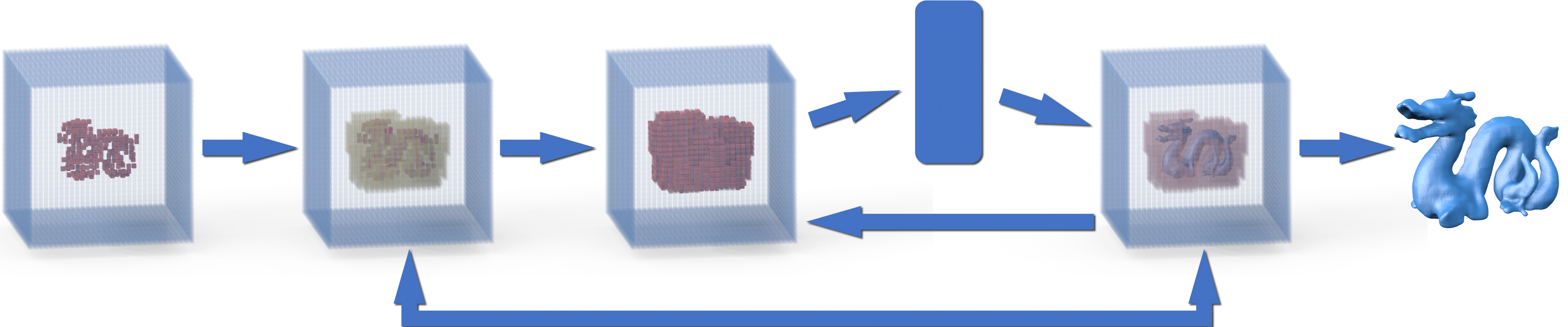}{
		\put(0.5,18.5){\small (a) partial scan}
		\put(19.5,18.5){\small (b) dilated domain}
		\put(39.5,18.5){\small (c) noise input}
		\put(56,21.5){\small (d) network}
		\put(60,15.5){$f$}
		\put(70.8,18.5){\small (e) prediction}
		\put(87.5,15.5){\small (f) extracted surface}
		\put(56.5,7.5){\small update domain}
		\put(44,1.5){\small minimize distance}
    }\end{overpic}
    \vspace{-5mm}
    \caption{Illustration of the pipeline. Given an incomplete shape in the form of truncated signed distance function (TSDF) (a), we build a dilated sparse volume with initial regions to be completed (b), and fill the sparsity pattern with noise (c) which is then fed to a 3D sparse encoder/decoder CNN (d) to recover a shape (e). The network is optimized to minimize the differences between (e) and (b) for the known shape regions only. The generation domain is updated gradually during the optimization iterations, to maintain the sparsity of the TSDF volume while also ensuring coverage of the missing regions. After sufficient iterations, we obtain the completed shape (f) extracted from the optimized TSDF. Fig.~\ref{fig:recon_process} shows the recovered shapes throughout the iterations. }
    \label{fig:pipeline}
    \vspace{-2mm}
\end{figure*}

We present a category-agnostic completion approach that uses deep neural networks to reconstruct single 3D shapes, avoiding the design of heuristic rules or the collection of large data sets.
Inspired by the works on image inpainting \cite{DeepImagePrior,deepDecoder} that complete a single image using the inherent \textit{deep prior} of CNNs, 
we train a sparse 3D CNN \cite{SparseConv2018} unsupervised on the single input shape to complete the missing regions (Fig.~\ref{fig:teaser}).
The pipeline of our framework is illustrated in Fig.~\ref{fig:pipeline}:
given an incomplete shape represented in the truncated signed distance function (TSDF) format commonly used for 3D scanning \cite{VolumeIntegration_SIG96}, 
we optimize a simple encoder/decoder network with stochastic gradient descent iterations to recover both the known shapes and missing regions from an input noise tensor, 
with the domain of regions to complete updated gradually in the process to ensure spatial sparseness for computational efficiency.
As is common for image processing, we also use multi-scale networks for progressive completion, where the shapes for different scales are encouraged to be consistent (Sec.~\ref{sec:design}).

While the deep prior has only been vaguely related with the recurring similarity of natural image patches \cite{DeepImagePrior,DoubleDIP}, 
we interpret its effectiveness from the \emph{neural tangent kernel} (NTK) perspective \cite{NTK_NIPS2018} and show that the CNN structure leads to plausible completion because of the closeness of different shape patches in the NTK feature space (Sec.~\ref{sec:ntk_interp}). 
This novel interpretation guides us to design more efficient network architectures and learning mechanisms for shape completion, 
including using deeper networks and smoothness of latent feature maps to enhance shape similarity, early stopping to avoid overfitting to details, and augmentation to enlarge the NTK space of possible shapes for completion (Sec.~\ref{sec:design}).
While designs like early stopping have been empirically used by deep prior for image inpainting \cite{DeepImagePrior}, our NTK perspective explains why they help and enables more systematic designs.

We perform extensive ablation studies to validate the design choices and demonstrate the NTK interpretation, and compare the proposed approach with both traditional and learning based completion methods (Sec.~\ref{sec:results}).
Our approach can adapt to 3D shapes of various styles and complete significant missing regions in a plausible way, while requiring no other training data than the incomplete shape itself.
Our method performs favorably to the related unsupervised 3D reconstruction methods that also use intrinsic regularities of deep neural networks \cite{hanocka2020point2mesh,Williams_2019_CVPR}.
Overall, our method provides an alternative approach to shape completion that is flexible and complements data-intensive supervised learning based methods.

To summarize, we make the following contributions:
\begin{itemize}
    \item A sparse 3D CNN based shape completion and reconstruction framework that fills large missing regions of a shape without relying on extra data sets for supervision.
    \item Interpretation of the CNN completion effectiveness, i.e. deep prior, in the framework of neural tangent kernels. Based on the NTK analysis, we propose more efficient designs for our completion framework, which are validated extensively by ablation tests.
    \item Comparison of our unsupervised shape completion with both traditional and deep learning based methods, which shows that our method recovers large missing regions more reasonably and adapts to the input shape more flexibly, complementing the supervised learning based methods. 
\end{itemize}
Code and data are publicly available to facilitate future research\footnote{URL: https://github.com/lei65537/NTKDeepPrior.}.

\begin{figure*}
    \centering
    \begin{overpic}[width=\linewidth]{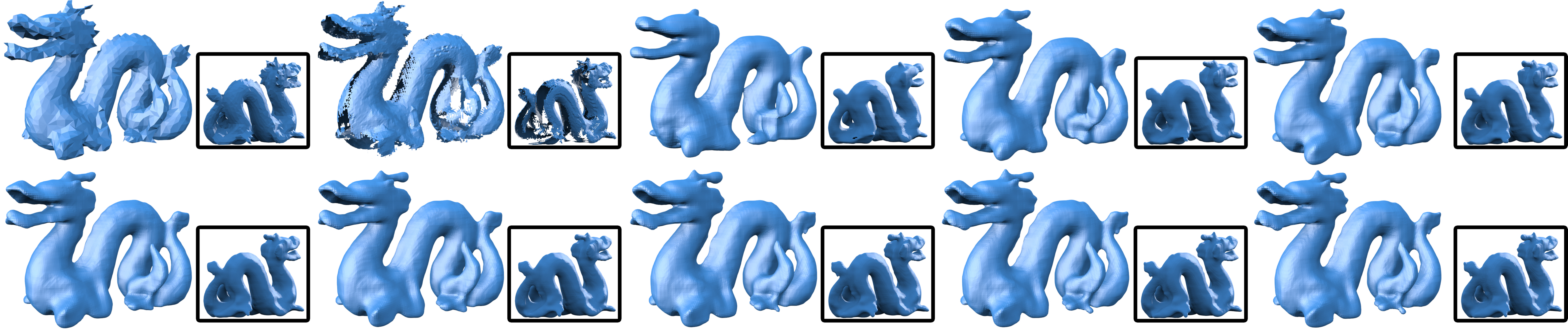}{
		\put(7,10){\small ground truth} 
		\put(27,10.3){\small input scan} 
		\put(47,10){\small iteration \#250} 
		\put(67,10){\small iteration \#500} 
		\put(87,10){\small iteration \#750} 
		\put(7,-0.8){\small iteration \#1000} 
		\put(27,-0.8){\small iteration \#1250} 
		\put(47,-0.8){\small iteration \#1500} 
		\put(67,-0.8){\small iteration \#1750} 
		\put(82,-0.8){\small iteration \#2000, final shape} 
    }\end{overpic}
    \vspace{-4mm}
    \caption{The evolution of predicted shapes by optimizing the 3D sparse CNN with stochastic gradient descent to fit to the incomplete input shape, where both the known shape and the unknown regions are gradually recovered. }
    \label{fig:recon_process}
    \vspace{-3mm}
\end{figure*}
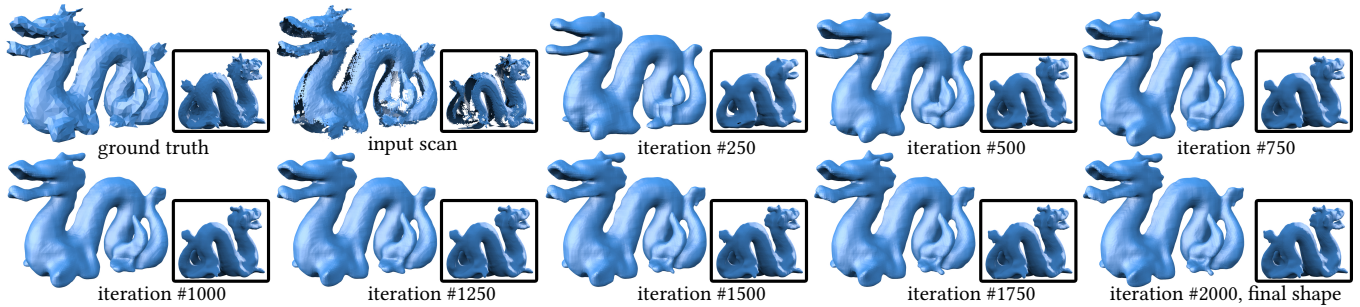

\section{Related work}

\paragraph{Shape completion and reconstruction.}

To complete the missing data is an inherently under-constrained problem; therefore, existing works use various kinds of priors to regularize the completion \cite{Berger2013,reconstar_eg14}.

Geometric smoothness represents a set of classic and widely used priors.
For example, on volumetric representations, the steady state diffusion prior fills holes with smooth interpolating surfaces, by solving Laplacian equations over signed distances \cite{DavisDiffusion2002} or Poisson equations over indicator values \cite{kazhdan2006poisson,kazhdan2013screened}.
Similarly on mesh representations, various kinds of smoothness functionals have been devised to optimize surface patches covering the holes so that they blend nicely with known regions \cite{ju2009fixing}, examples including Laplacian fairing \cite{LiepaHoleFilling}, biharmonic fairing \cite{zhao2007robust}, Willmore flow \cite{WillmoreFlow2004} and other PDEs \cite{XuFEM2009}.
The smoothness priors work robustly across shape categories, but generally handle small missing regions and do not adapt to shape features, as they are only aware of the proximate boundary conditions.

Geometric priors specially designed for particular kinds of shapes also exist; they abstract the incomplete shape with compact representations and then recover the regularized complete shape.
Examples include curve skeletons for generalized cylindrical shapes \cite{li2010analysis,Huang2013} and primitives (planes, cylinders, etc.) for CAD models \cite{schnabel2009completion}.
Symmetry and repetition of structures are also strong priors leading to high quality completion that respects these global relationships \cite{pauly2008discovering,sipiran2014approximate,speciale2016symmetry,thrun2005shape,facade2011}.
However, such hand-crafted symmetry rules can only be applied on shapes strictly obeying them.

Motivated by image inpainting and texture synthesis, a series of context-based methods \cite{context04,Bendels05,Harary:2014:CCS} try to synthesize the missing patches by copying from similar patches in the existing regions; 
the key problem therefore is to evaluate the similarity of patches.
Sharf et al.~\shortcite{context04} use fitted algebraic equations to represent the patch geometry, 
Bendels et al.~\shortcite{Bendels05} use regularly sampled geodesic fans, 
and Harary et al.~\shortcite{Harary:2014:CCS} use statistics of heat kernel signatures.
After evaluating patch similarity, these methods fill in the missing regions by pasting appropriate patches in a coarse-to-fine hierarchical manner.

Compared with these hand crafted priors, our approach is automatically aware of the inherent self-similarities of the incomplete shape.
While conceptually similar to the context based methods, 
our method encodes the patches as deep network feature maps with similarities defined by NTK, 
which is learned from the shape adaptively and inherently multi-scale due to the deep CNN structure.

Early data-driven priors try to explicitly search for the matching templates for incomplete shapes, and deform and assemble the retrieved templates appropriately for reconstruction \cite{pauly2005example,template05,shen2012structure,sung2015data,rock2015completing}.
With the development of deep learning, now an entire database of shapes can be encoded by deep neural networks, thus enabling shape completion by inferring the missing regions according to the shape dataset prior encoded by the neural network and conditioned on the incomplete input \cite{wu20153d,3depn,HanComplete2017,firman2016structured,fan2017point,rezende2016unsupervised,riegler2017octnetfusion,sharma2016vconv,smith2017improved,Stutz2018CVPR,DeepSDF,dai2020sgnn}.
In particular, a flurry of MLP parameterized implicit function based representations show strong competency in generating high quality 3D reconstructions \cite{DeepSDF,chibane2020implicit,Chabra2020LocalSDF,Jiang2020LocalScene,Peng2020ConvOcc,atzmon2020sal,gropp2020icml}.
By using the strong database prior, these powerful data driven approaches can recover shapes from very little input data.
However, while in large scale, the databases organized semantically by shape categories can still be limited in covering all shape variations and combinations, restricting the generalization of data driven methods to unseen shapes.

Instead of relying on category-specific priors encoded by large databases, \cite{Williams_2019_CVPR,hanocka2020point2mesh} are two recent works that use the inherent regularities of deep neural networks to reconstruct 3D shapes from point clouds, with good robustness in handling noisy point clouds without reliable normal data and insufficiently sampled regions with repetitive geometric textures.
For point cloud denoising, there is also work by \cite{hermosilla2019totaldenoise} that adopts unsupervised learning regularized by the prior of spatial and color closeness of measured points.

Our method is in the same vein, as it completes by unsupervised learning on a single input shape, to detect and apply patterns in a self-coherent way, and does it implicitly by stochastic gradient descent (SGD) optimization of a 3D CNN. 
While incapable of handling extreme missing data due to the lack of database prior, 
our approach is more flexible in handling new shape variations, therefore complementing the database driven approaches (Sec.~\ref{sec:comparison}).

\vspace{-2mm}
\paragraph{Deep priors.}
Like shape completion, many image recovery tasks, e.g. denoising, super resolution and inpainting from a corrupted input image, are ill-posed and typically formulated as a regularized data fitting problem, where hand-crafted regularization priors like total variation are used.
Ulyanov et al. \shortcite{DeepImagePrior} show that, instead of using these explicit regularization terms, by parameterizing the recovered image as an output of a CNN and training the CNN on the data fitting problem, the recovered image is automatically plausible and free from artifacts;
the inherent regularization ability by a CNN is therefore called the deep image prior.
In addition, through extensive experiments the authors have noted several characteristics of the deep image prior, e.g. that a randomly initialized CNN quickly learns signals but shows impedance at picking up noise, and that the output of a CNN shows self-similarities among image patches, but make no further explanations as to how these happened.

Subsequent works have studied the deep image prior for different network structures \cite{deepDecoder}, used the deep prior for foreground and background segmentation due to their different statistics \cite{DoubleDIP}, and for regularizing the inpainting of depth images for multiview stereo \cite{ghosh2020depthPrior}.
Notably, Gadelha et al.~\shortcite{Gadelha_2019_ICCV} use differentiable projection operations and deep prior to reconstruct 3D shapes from silhouettes and depth images by optimizing a dense 3D occupancy CNN, which is computationally more expensive than our sparse CNN approach and does not provide an analysis of the deep prior or guided network designs as we do.
Different from the deep prior of a CNN, Williams et al.~\shortcite{Williams_2019_CVPR} propose the notion of deep geometric prior for reconstructing surface patches parameterized by MLPs from point clouds, where the regularity of the surface patches is regarded as the prior inherent to MLP networks.
Hanocka et al.~\shortcite{hanocka2020point2mesh} rely on the notion of self-prior to detect and apply repetitive patterns on surface meshes for shape reconstruction,
where they empirically observe that the shared convolution kernels of a mesh CNN~\cite{hanocka2019meshcnn} would encourage self-similarity, but remark that a deeper understanding for the self-prior is lacking.

In this paper, we extend the deep prior to 3D sparse CNNs and use its inherent regularity to complete 3D shapes. 
Moreover, we take the neural tangent kernel perspective and give an interpretation of the deep prior, which explains many empirical observations in \cite{DeepImagePrior} and guides us to design more efficient networks and learning mechanisms for shape completion.

Concurrent to our work, Tachella et al.~\shortcite{tachella2020NTDenoiser} relate image CNNs to their non-local denoising feature also observed in the deep prior through the NTK analysis. 
In addition, Cheng et al.~\shortcite{Cheng_2019_CVPR} interpret the deep image prior by a novel spatial Gaussian process that holds asymptotically on stationary input as the network width goes to infinity, and adopt the noise-augmented stochastic gradient Langevin dynamics \cite{SGLD2011} to avoid overfitting. 
In comparison, the tangent kernel analysis we take holds even for finite network width and we show in detail how it guides network design for effective shape completion.

\section{Preliminary on Neural Tangent Kernel}
\label{sec:ntk_background}

Neural tangent kernel (NTK) \cite{NTK_NIPS2018} uses the kernel method \cite{shawe-taylor_cristianini_2004} to analyze the gradient descent based learning dynamics of a neural network.
In the NTK perspective, it is shown that the update of network parameters follows the path of kernel regression gradient descent, therefore establishing a precise model showing that the trained network is equivalent to a kernel regression function.
We briefly review the relevant results below.

Let $f(\vb*{\theta},\cdot)$ be a network parameterized by $\vb*{\theta}$, $\{\vb*{x}_i \in \mathcal{X}|i\in[n]\}$  the set of training samples with ground truth labels $\vb*{u}=\left(u_i\in\mathbb{R}\right)_{i\in[n]}$, and $\vb*{y} = \left(f(\vb*{\theta},\vb*{x}_i)\right)_{i\in[n]}$ the network output on the samples. 
Suppose the squared training loss $\ell(\vb*{\theta}) = \frac{1}{2}||\vb*{u} - \vb*{y}||^2$ is used.
Minimizing $\ell(\vb*{\theta})$ by gradient descent with infinitesimal learning rate gives $\dv{\vb*{\theta}}{t} = -\nabla\ell\left(\vb*{\theta}(t)\right)$, where $t$ is the time dimension of the dynamics.
On the other hand, we can view the update as searching for the optimal function $f^* = \argmin_{f\in\mathcal{F}}{\ell}$ in the function space $\mathcal{F} = \{f(\vb*{\theta},\cdot) : \mathcal{X}\rightarrow \mathbb{R}\}$ defined by the network structure and parameterized by $\vb*{\theta}$.
Indeed, by substituting the gradient dynamics of $\vb*{\theta}$ into $\dv{\vb*{y}}{t} = \pdv{\vb*{y}}{\vb*{\theta}}\cdot\dv{\vb*{\theta}}{t}$ (see Appendix~\ref{sec:appendix:ntk_derivation} for details), we obtain the kernel gradient dynamics for solving the learning problem as kernel regression:
\begin{equation}
\dv{\vb*{y}}{t} = \vb{K}(t)\cdot\left(\vb*{u} - \vb*{y}(t)\right),
\label{eq:kernel_gradient}
\end{equation}
where $\vb{K}\in \mathbb{R}^{n\times n}$ is the kernel Gram matrix with each entry $\vb{K}_{ij} = \textrm{ker}(\vb*{x}_i, \vb*{x}_j)$
evaluating the tangent kernel function $\textrm{ker}(\cdot,\cdot) : \mathcal{X}\times \mathcal{X}\rightarrow \mathbb{R}$ between a pair of data samples.
In particular, 
\begin{equation}
\textrm{ker}(\vb*{x}_i,\vb*{x}_j) = \left<\phi(\vb*{x}_i),\phi(\vb*{x}_j)\right>,\quad \phi(\vb*{x}) = \pdv{f(\vb*{\theta},\vb*{x})}{\vb*{\theta}}
\label{eq:kernel_function}
\end{equation}
is the tangent kernel feature map.
The kernel feature map encodes the network structure through chained differentiation;
in the context of CNNs, it is closely related to the commonly known layer-wise feature maps (Sec.~\ref{sec:ntk_interp}).

The neural tangent kernel defines its corresponding reproducing kernel Hilbert space (RKHS) $\mathcal{H} = \overline{\mathrm{span}\{\textrm{ker}(\cdot,\vb*{x})|\vb*{x}\in\mathcal{X}\}}$, 
as the closure of the vector space spanned by the kernel function partially applied on the input  samples \cite{Aronszajn_RKHS,scholkopf2001generalized}.
Moreover, due to the representer theorem \cite{scholkopf2001generalized}, assuming convergence of the gradient descent training process as $t{\rightarrow}{+}\infty$, the fully trained network with parameters $\vb*{\theta}^*$ is equivalent to the kernel regression predictor that resides in $\mathcal{H}$:
\begin{equation}
f(\vb*{\theta}^*,\vb*{x}) = \left(\textrm{ker}(\vb*{x}, \vb*{x}_i)\right)_{i\in[n]} \vb{K}^{-1} \vb*{u}, \quad \forall \vb*{x} \in \{\vb*{x}_i\}.
\label{eq:kernel_regression}
\end{equation}
The equation describes the generalization of neural networks \cite{NTK_NIPS2018}: for a data sample from the training data distribution, 
the output of the learned neural network is the kernel weighted combination of labels for the training data samples.

It is shown that for randomly initialized and infinitely wide neural networks the kernel remains constant without learning from data in the training process \cite{NTK_NIPS2018,CNTK_NIPS2019,Lee_NIPS2019}, 
thus allowing training convergence analysis and constructing the NTK equivalent of infinite-width neural networks algorithmically \cite{GNTK_NIPS2019,neuraltangents2020}.
Meanwhile, Hanin and Nica~\shortcite{Hanin2020Finite} show that in the finite width case the kernel matrix adapts to data during training.

In this paper, we study the implications of the results in Eqs.~(\ref{eq:kernel_gradient})-(\ref{eq:kernel_regression}) for a finite-width CNN trained on and therefore adapted by a single incomplete shape.
In general for a CNN, 
because the NTK feature map $\phi(\vb*{x})$ corresponds to the compounded feature maps evaluated for $\vb*{x}$ at different layers of the network,
for two patches $\vb*{x}_1,\vb*{x}_2$ whose feature maps are more similar, 
they have larger tangent kernel correlation $\mathrm{ker}(\vb*{x}_1,\vb*{x}_2)$ and more similar network outputs (Sec.~\ref{sec:ntk_interp}).
This observation provides the foundation for interpreting the effectiveness of deep prior; our use of deep prior is thus an example of low-shot learning on the few known regions of an incomplete shape \cite{Mu2020Gradients}.

\section{Method}
\label{sec:method}
We formulate the task of deep prior based shape completion in Sec.~\ref{sec:formulation}, apply the NTK analysis to deep prior to reveal its effectiveness in Sec.~\ref{sec:ntk_interp} and propose NTK informed designs for shape completion in Sec.~\ref{sec:design}.

\subsection{Shape completion via deep prior}
\label{sec:formulation}

Assuming the objects are normalized into the unit cube centered at the origin in $\mathbb{R}^3$,
we use a regular grid of dimension $H\times H \times H$ to discretize the cubic domain.
An incomplete 3D shape is represented as a TSDF discretized on the grid points, i.e., $\vb*{u}_{i,j,k} \in [-1,1]$ for any grid point $(i,j,k)\in \Omega$, where $\Omega$ is a narrow band of grid points that have known signed distance values to the observed object boundary and the truncated signed distances have been normalized to the range $[-1,1]$.
We further assume the domain where shapes can potentially be recovered as $\mathcal{M} \supset \Omega$, and a corresponding mask vector $\vb*{m}$ such that $\vb*{m}_{i,j,k}=1$ for every known grid point $(i,j,k)\in \Omega$ and $\vb*{m}_{i,j,k}=0$ for grid points $(i,j,k)\in \mathcal{M}{\setminus}\Omega$ that are unknown but potentially contain missing shapes.
The domain $\mathcal{M}$ is obtained by dilating $\Omega$; we present the details later in Sec.~\ref{sec:implementation}.

Following the deep image prior \cite{DeepImagePrior} methodology,
we parameterize the recovered shape TSDF by a 3D sparse CNN $f(\vb*{\theta}, \vb{z}) : (i,j,k) \in \mathcal{M} \rightarrow \mathbb{R}$ that is defined on the domain $\mathcal{M}$ only \cite{SparseConv2018}, 
where $\vb*{\theta}$ are the trainable parameters, and $\vb{z}$ is the input vector defined on $\mathcal{M}$ that assigns random noise independently sampled from the uniform distribution $U(0,0.1)$ to each grid point.
We formulate completion as a fitting problem that tries to recover the known shape:
\begin{equation}
\vb*{\theta}^*=\argmin_{\vb*{\theta}} \left\|\vb*{m}\odot\left(\text{clp}_\eta\left(f(\vb*{\theta},\vb{z})\right) - \text{clp}_\eta(\vb*{u})\right)\right\|^2,\ \ \vb*{y}^*=f(\vb*{\theta}^*,\vb{z}),
\label{eq:tsdf_min}
\end{equation}
where $\odot$ is the component-wise product and the clipping function 
\begin{equation}
\text{clp}_\eta(x)=\min(\eta,\max(-\eta,x))
\label{eq:clip_function}
\end{equation}
focuses the network on regressing the distance function gradation close to the zero level set surface, with $\eta=0.5$ in our experiments.
The boundary surface of the completed shape can then be extracted from $\vb*{y}^*$ by Marching cubes \cite{lorensen1987marching}.

Before introducing the details of the sparse CNN and its training in Sec.~\ref{sec:design}, 
we first relate the deep prior exhibited by such a regression to NTK in the next section, 
which in turn guides our design for more effective shape completion.

\subsection{Deep prior from the NTK perspective}
\label{sec:ntk_interp}

The NTK perspective when applied to corresponding networks reveals their structural inductive biases.
By analyzing how the surface patches are embedded and related through the NTK of a CNN, we show that the patch self-similarity as observed in the deep prior is naturally induced.

For notation simplicity, we consider a 1D fully convolutional neural network with $L$ layers of convolutions as the example; but the analysis can be readily extended to the 3D CNNs used in our completion task. 

Denote the feature map width (or channel size) for each layer as $\{C^{(h)}|h\in[L]\}$, the feature map spatial size as $\{P^{(h)}\}$ and the feature maps as $\{\vb*{x}^{(h)}{\in}\mathbb{R}^{P^{(h)}{\times}C^{(h)}}\}$.
Therefore, $\vb*{x}^{(0)}$ is the network input and $\vb*{y} = \vb*{x}^{(L)}$ is the output.
Further denote the trainable parameters of convolution operators at the $h$-th layer as $\vb*{\theta}^{(h)}\in\mathbb{R}^{Q^{(h)}{\times}C^{(h-1)}{\times}C^{(h)}}$, 
with the filter spatial size $Q^{(h)}$ being an odd number for simplicity, and the bias parameters as  $\vb*{b}^{(h)}\in\mathbb{R}^{C^{(h)}}$.
At each layer, we have two operations: the convolution and bias addition $\widetilde{\vb*{x}}^{(h)} = \vb*{\theta}^{(h)}*\vb*{x}^{(h-1)} + \vb{1}_{P^{(h)}}\otimes\vb*{b}^{(h)}$, where $*$ is the convolution operator and $\vb{1}_{P^{(h)}}$ is a tensor of shape ${P^{(h)}}$ with constant one elements, followed by  $\vb*{x}^{(h)} = \sigma(\widetilde{\vb*{x}}^{(h)})$ which consists of operations without learnable parameters, like nonlinear activation, normalization, etc.

We further define a sparse structure tensor $\vb{S}^{(h)}$ that encodes the linear convolution operation for layer $h$, so that the gradients can be more compactly expressed.
In particular, $\vb{S}^{(h)}$ of shape ${P^{(h-1)}}{\times}C^{(h-1)}{\times}{Q^{(h)}}{\times}C^{(h-1)}{\times}C^{(h)}{\times}{P^{(h)}}{\times}C^{(h)}$ is defined as:
\begin{equation}
    \vb{S}^{(h)}_{ic_1,jc_2c_3,kc_4} = 
    \begin{cases} 
    1, & \textrm{if\ } i = j-\frac{Q^{(h)}-1}{2} + k, c_1=c_2, c_3=c_4\\
    0, & \textrm{otherwise.}
    \end{cases}
    \label{eq:structure_tensor}
\end{equation}
The forward pass can then be defined through tensor contraction and addition as $\widetilde{\vb*{x}}^{(h)} = \vb{S}^{(h)}\left(\vb*{x}^{(h-1)}, \vb*{\theta}^{(h)}, \cdot\right) + \vb{1}_{P^{(h)}}\otimes\vb*{b}^{(h)}$.
Note that zero padding is automatically implied by the structure tensor.
Meanwhile, the partial derivatives used for the backward pass are defined by recursion:
\begin{align}
    \pdv{f}{\vb*{\theta}^{(h)}} &= \vb{S}^{(h)}\left(\vb*{x}^{(h-1)}, \cdot, \pdv{f}{\widetilde{\vb*{x}}^{(h)}} \right), \\
    \pdv{f}{\vb*{x}^{(h-1)}} &= \vb{S}^{(h)}\left(\cdot, \vb*{\theta}^{(h)}, \pdv{f}{\widetilde{\vb*{x}}^{(h)}} \right),   \label{eq:partial_f_theta}\\
    \pdv{f}{\vb*{b}^{(h)}} &= \pdv{f}{\widetilde{\vb*{x}}^{(h)}}\left(\vb{1}_{P^{(h)}},\cdot\right).
\end{align}
It is obvious that the NTK features $\frac{\partial f}{\partial \vb*{\theta}^{(h)}}$ and $\frac{\partial f}{\partial \vb*{b}^{(h)}}$ (Eq.~\ref{eq:kernel_function}) depend on the feature maps of two neighboring layers only. 
In addition, by adding more layers into consideration, such a structure implies that as the layer depth goes deeper in the backward direction, the partial derivatives become more similar for two patches of the output, which means their NTK correlation becomes larger and their outputs given by the trained network more similar.

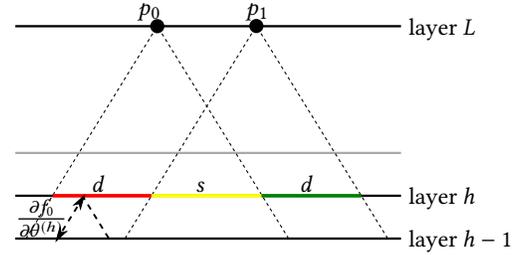
\begin{figure}
	\centering
	\begin{overpic}[width=0.6\linewidth]{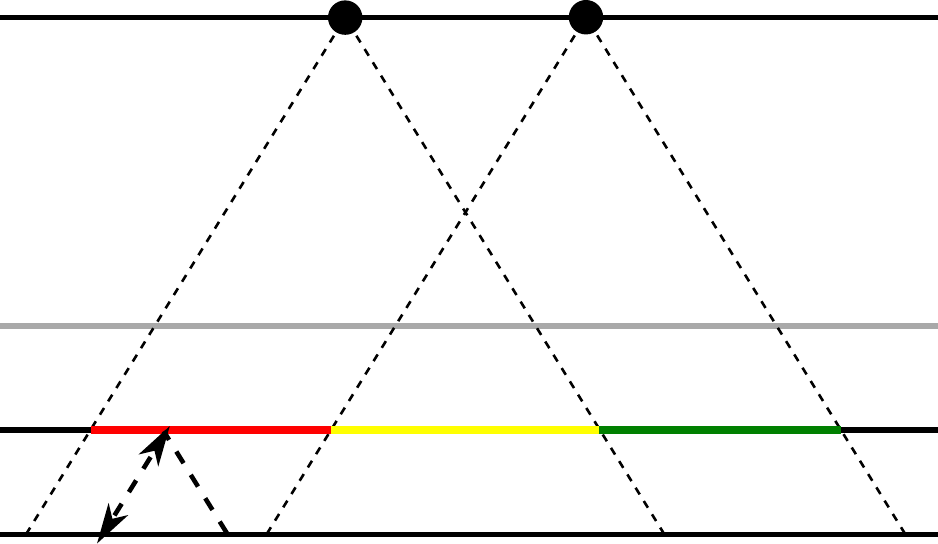}{
			\put(102,-1){layer $h-1$}
			\put(102,10){layer $h$}
			\put(102,54){layer $L$}
			\put(32,59){$p_0$}
			\put(60,59){$p_1$}
			\put(20,13){$d$}
			\put(47,13){$s$}
			\put(74,13){$d$}
			\put(0,4.5){$\pdv{f_0}{\theta^{(h)}}$}
	}\end{overpic}
	\vspace{-1mm}
	\caption{Illustration of the tangent kernel similarity due to overlapping receptive fields for two output points. 
		Since $\pdv{f}{\theta^{(h)}}$ depends on the two neighboring feature maps at layers $h-1$ and $h$ only (Eq.~\ref{eq:partial_f_theta}), the shared region at layer $h$ induces the same contribution to $\pdv{f_0}{\theta^{(h)}}$ and $\pdv{f_1}{\theta^{(h)}}$, leading to similarity in NTK between the two output points.
	}
	\label{fig:overlap}
	\vspace{-3mm}
\end{figure}

The situation is illustrated graphically in Fig.~\ref{fig:overlap}.
In the figure, the shared region which induces the same contribution to the tangent kernel features for two output points $p_0,p_1$ is marked in yellow and has length $s$, while the non-overlapping regions are marked in red and green respectively and have equal length $d$.
While $d$ as the distance between the two points is fixed, $s$ gets bigger as we move backward to earlier layers.
Therefore, the similarity between $\pdv{f_0}{\theta^{(h)}}$ and $\pdv{f_1}{\theta^{(h)}}$ gets larger for decreased $h$.
In addition, when the cone of receptive field expands out of the boundary of the feature maps, the non-overlapping regions begin to become similar due to the zero padding, which again contributes to the similarity of the tangent kernel features for the two points.
Note that for $\frac{\partial f}{\partial \vb*{b}^{(h)}}$, the dimensions of the tangent kernel feature map corresponding to the bias parameters, they are the same for all patches.
Greater similarity in the tangent kernel feature in turn means closer outputs by the trained network (Eq.~\ref{eq:kernel_regression}).
Echoing how geometric shapes are globally coherent but locally varying, this CNN structural property explains why over the missing regions the trained network produces shapes that resemble and blend well with the known regions throughout the scales, as captured by the deep prior.

\begin{figure}
	\centering
	\begin{subfigure}[b]{0.23\textwidth}
		\centering
		\includegraphics[width=\textwidth]{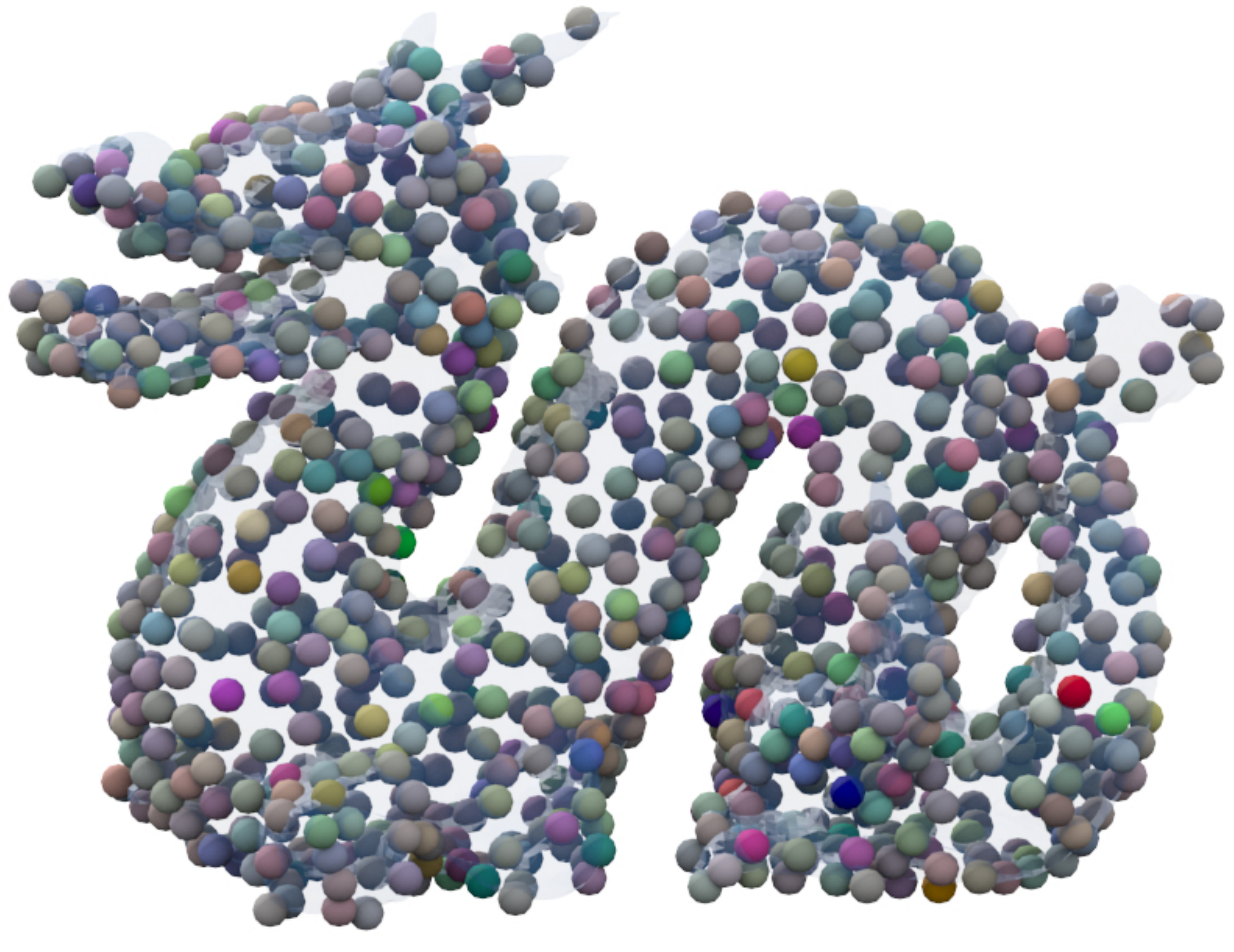}
		\caption{}
		\label{fig:NTK_before_vis}
	\end{subfigure}
	\hfill
	\begin{subfigure}[b]{0.23\textwidth}
		\centering
		\includegraphics[width=\textwidth]{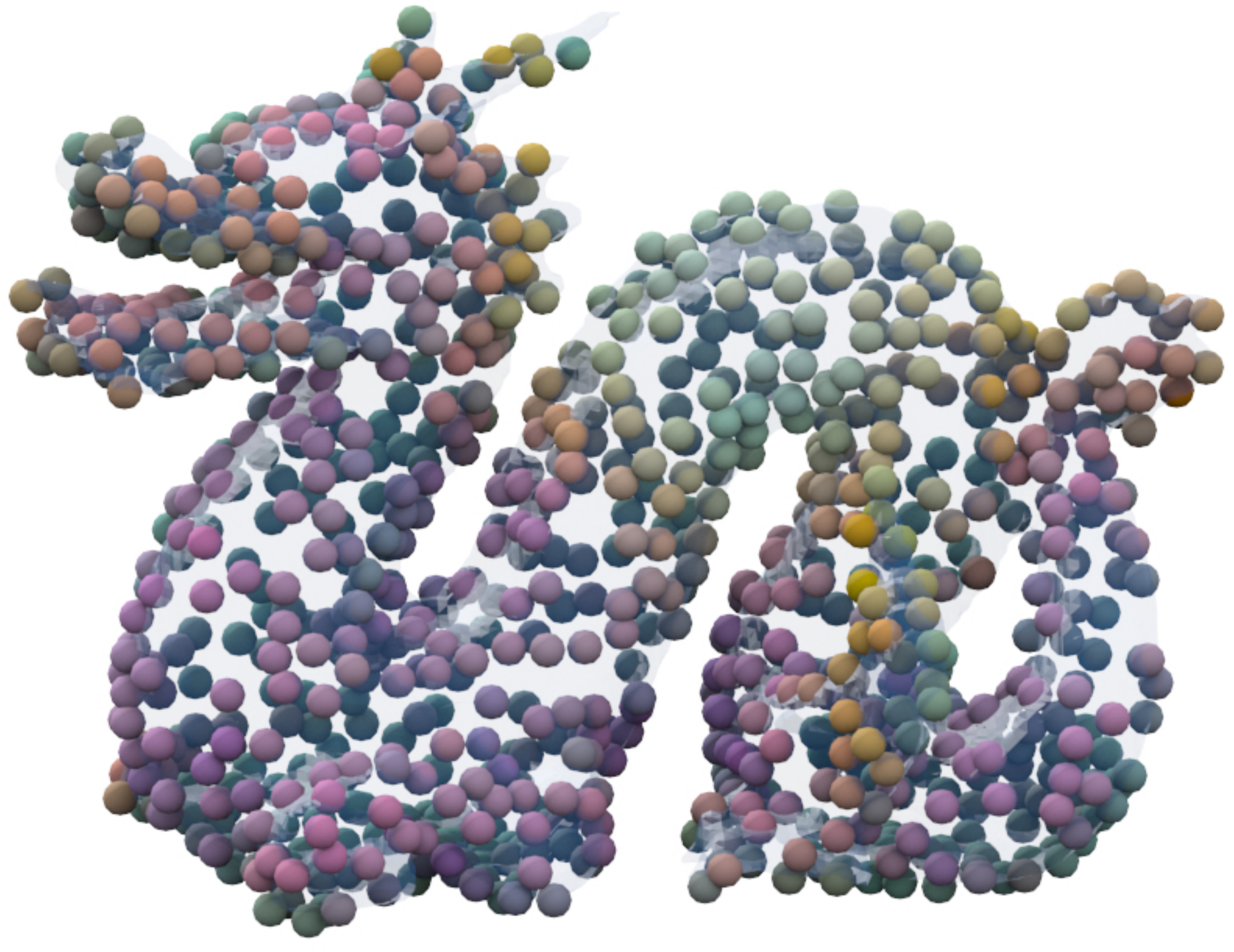}
		\caption{}
		\label{fig:NTK_after_vis}
	\end{subfigure}
	\hfill
	\begin{subfigure}[b]{0.23\textwidth}
		\centering
		\includegraphics[width=\textwidth]{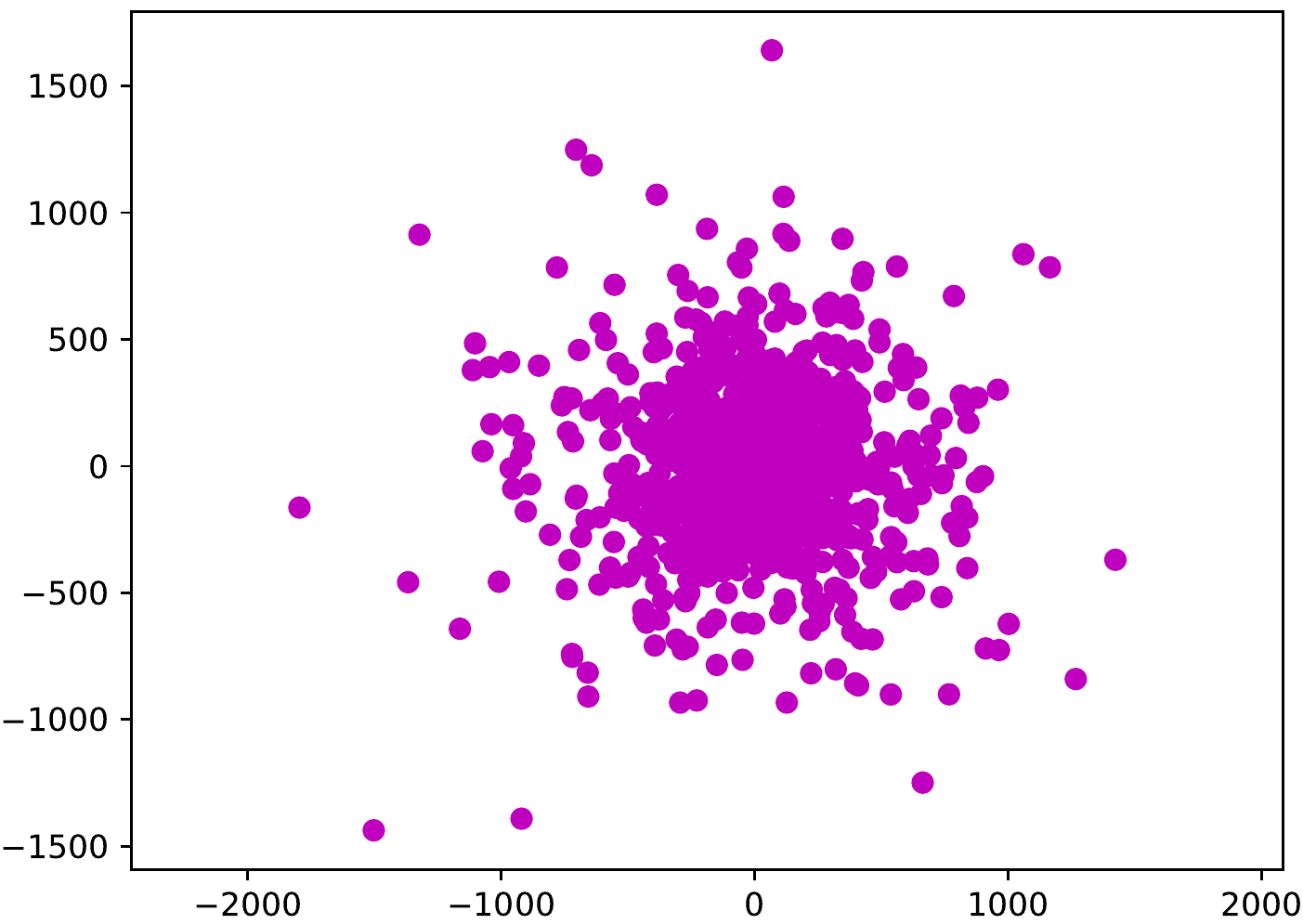}
		\caption{}
		\label{fig:NTK_before_scatter}
	\end{subfigure}
	\hfill
	\begin{subfigure}[b]{0.218\textwidth}
		\centering
		\includegraphics[width=\textwidth]{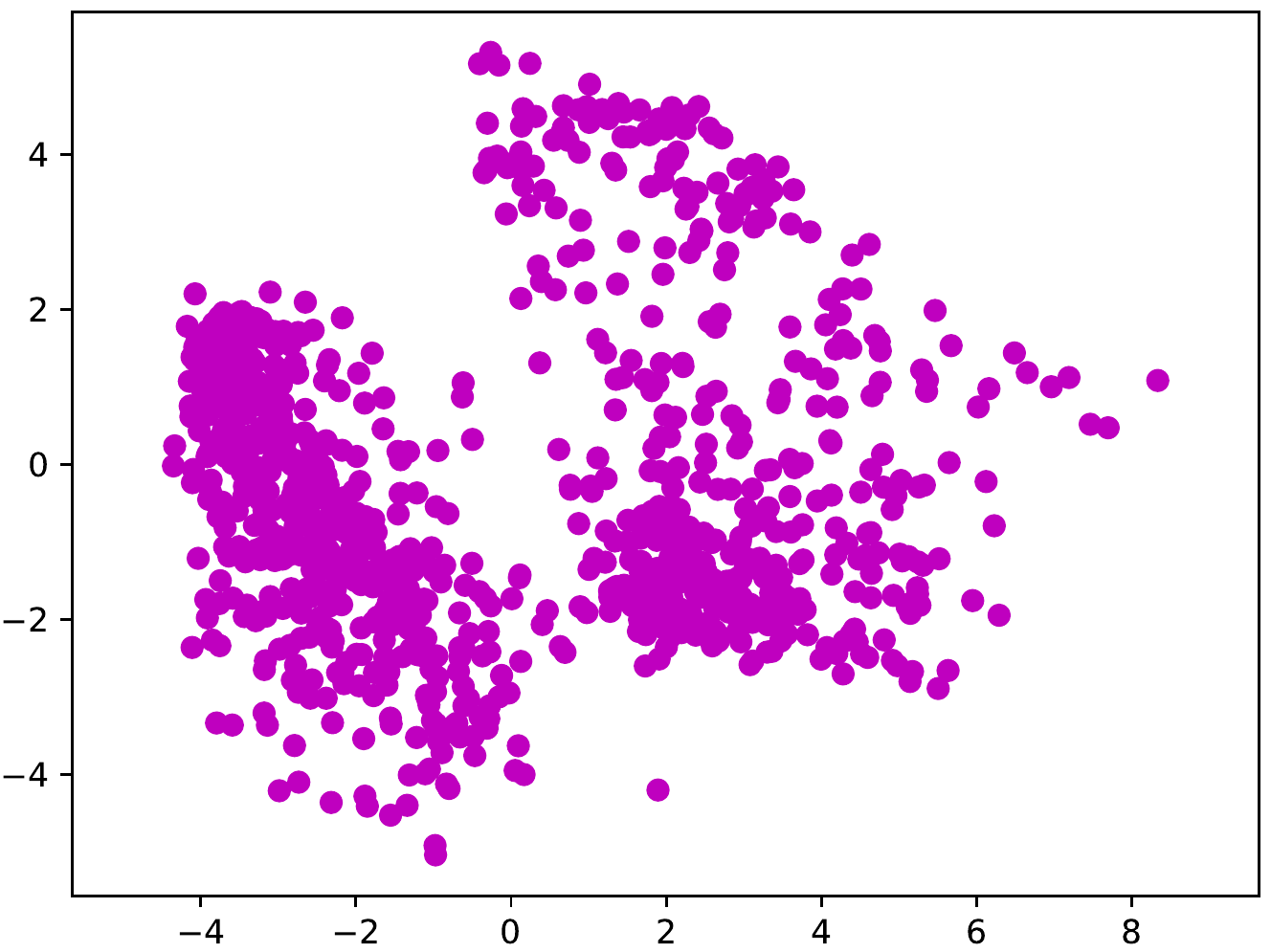}
		\caption{}
		\label{fig:NTK_after_scatter}
	\end{subfigure}
    \vspace{-3mm}
	\caption{Kernel PCA analysis with the NTK, before (left) and after (right) network training. The top three principal component coordinates are used for color coding the spatial points in (a),(b), and the top two principal component coordinates used for 2D plots in (c),(d). Before training, the points are embedded randomly in NTK, but regular patterns emerge with self similarities after training.}
	\label{fig:NTK_vis_before_after_opt}
	\vspace{-3mm}
\end{figure}

The phenomenon can also be confirmed by directly visualizing the kernel distances between sampled points on the 3D surfaces.
In particular, we sample 1k random points uniformly from the dragon model, and apply kernel PCA \cite{kernelPCA} to visualize their similarities under NTK, as shown in Fig.~\ref{fig:NTK_vis_before_after_opt}.
The kernel PCA takes their Gram matrix as specified in Eq.~\ref{eq:kernel_function} as input and assigns to each sample point the top three principal component coordinates, which are used as RGB color channels\footnote{The coordinates are first truncated by the $3\sigma$ rule to avoid outlier dominance, where $\sigma$ is the standard deviation of the top component coordinates, and then normalized uniformly and shifted into the range $[0,1]$ to fit into the RGB color space.} in Fig.~\ref{fig:NTK_vis_before_after_opt}(a),(b), and the top two component coordinates for scattered plot in Fig.~\ref{fig:NTK_vis_before_after_opt}(c),(d).
We can see that while before training, the NTK embeddings of the sample points are randomly distributed close to a normal distribution (Fig.~\ref{fig:NTK_vis_before_after_opt}(c)) and there is little spatial smoothness over the surface (Fig.~\ref{fig:NTK_vis_before_after_opt}(a)), after training, the spatial smoothness is very obvious, with similar structures showing closeness in NTK (see the two separated points on the two horns in Fig.~\ref{fig:NTK_vis_before_after_opt}(b)), and the 2D projection (Fig.~\ref{fig:NTK_vis_before_after_opt}(d)) shows regular clustering patterns than randomness.

\subsection{The completion network}
\label{sec:design}

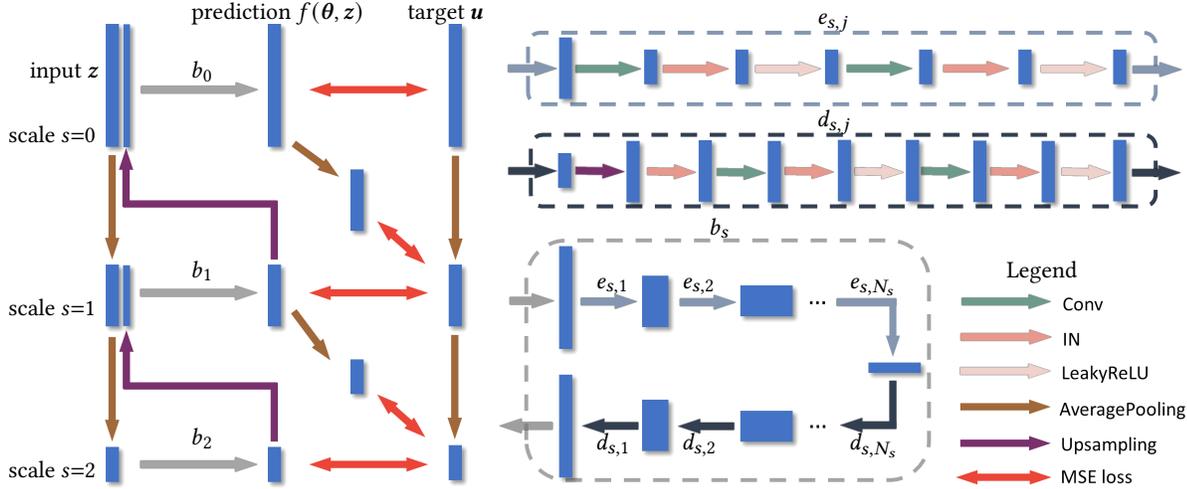
\begin{figure*}
    \centering
    \begin{overpic}[width=0.8\textwidth]{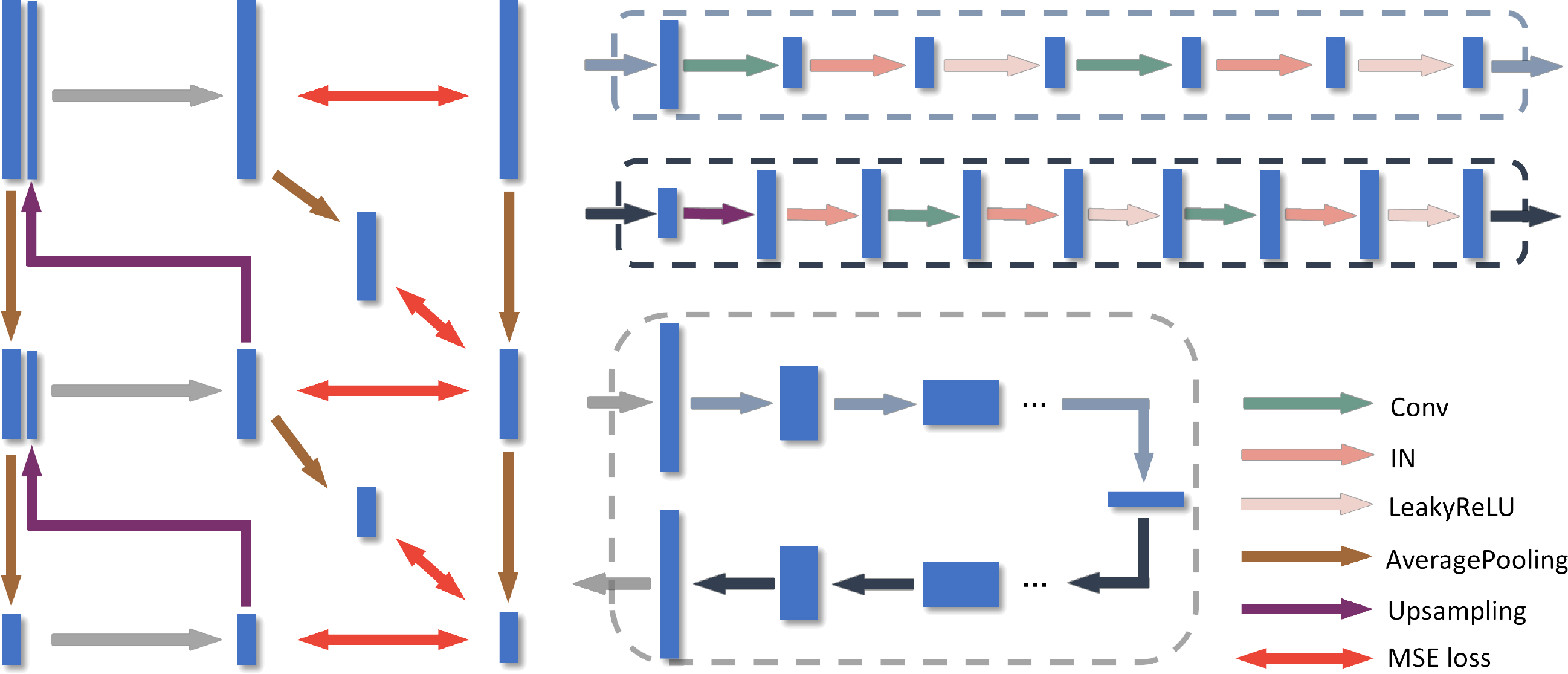}{
        \put(-7,38){input $\vb*{z}$}
        \put(8,43.5){prediction $f(\vb*{\theta},\vb*{z})$}
        \put(28,43.5){target $\vb*{u}$}
        \put(-9,32){scale $s{=}0$}
        \put(-9,16){scale $s{=}1$}        
        \put(-9,1){scale $s{=}2$}
        \put(8,38){$b_0$}
        \put(8,19.5){$b_1$}
        \put(8,4){$b_2$}
        \put(56,23.5){$b_s$}
        \put(66,43){$e_{s,j}$}
        \put(66,33.5){$d_{s,j}$}
        \put(83.5,19.5){Legend}
        \put(45.5,3.5){$d_{s,1}$}
        \put(45.5,18.5){$e_{s,1}$}
        \put(53.5,3.5){$d_{s,2}$}
        \put(53.5,18.5){$e_{s,2}$}
        \put(69,3.5){$d_{s,N_s}$}
        \put(69,18.5){$e_{s,N_s}$}
    }
    \end{overpic}
    \vspace{-2mm}
    \caption{The three-scale network structure. Each scale $s$ uses the downsampled noise vector $\vb*{z}\downarrow_{s}$ as input and the downsampled target TSDF $\vb*{u}\downarrow_{s}$ as supervision, mapped through the base block network $b_{s}$ that is a symmetric pair of encoder $[e_{s,j}]$ and decoder $[d_{s,j}]$ without skip connections. $e_{s,j}$ contains a convolution with stride 2 for downsampling followed by another convolution. $d_{s,j}$ contains nearest neighbor upsampling followed by two convolutions. All convolutions are followed by instance normalization and leaky ReLU activation. More details about the network structure are given in Appendix \ref{appdx:network_architecture}. }
    \label{fig:netstruct}
    \vspace{-3mm}
\end{figure*}

We design our effective 3D CNN for completion by drawing on insights offered from the NTK perspective.
Generally, we make the following design choices:
\begin{enumerate}[(a)]
    \item We use CNNs that enable the sharing of deep features for different patches to induce similarity between patches of known regions and patches to complete.
    \item We further enhance the smoothness of generated patches by Laplacian regularization of feature maps close to the final \mbox{output}.
    \item We use early stopping for training to make sure that the missing regions capture the dominant features of the shape but avoid focusing on extra details that harm shape regularity.
    \item We augment the shape by randomly rotating it around its center and train the network on the augmented set, therefore using the known patches with different poses to form the NTK function space (Sec.~\ref{sec:ntk_background}) and completing the missing regions with more flexible choices.
\end{enumerate}
We ablate and discuss the above design choices extensively in Sec.~\ref{sec:ablation}.
The first two designs have further details as presented next.

\paragraph{Multi-scale hierarchy.}
It is common practice to generate high quality images in a coarse-to-fine multi-scale manner. 
We also use a three-scale hierarchical network, where the base block in each scale is an encoder-decoder, to achieve completion of both large scale structures and small details (Fig.~\ref{fig:netstruct}).
From the NTK perspective, both the encoder-decoder structure and the use of coarse scale sub-networks enlarge the overlapping of deep features for the final fine scale output patches, thus allowing recovering more coherent structures globally.

At each scale, the base block network is additionally supervised to minimize the fitting loss (Eq.~\ref{eq:tsdf_min}) to a down-sampled version of the incomplete input shape, defined as
\begin{equation}
\ell_{fit}^{(s)} = \left\|\vb*{m}\downarrow_{s}\odot\left(\text{clp}_\eta\left(f^{(s)}\left(\vb*{\theta},\vb{z}\downarrow_{s};f^{(s+1)}\uparrow_1\right)\right) - \text{clp}_\eta(\vb*{u}\downarrow_{s})\right)\right\|^2, 
\end{equation} 
for $s=0,1,2$, ranging from fine to coarse scales.
$f^{(s)}$ is the base block output at scale $s$, which directly takes as input the concatenation of downsampled noise vector $\vb{z}\downarrow_{s}$ with the upsampled coarse output $f^{(s+1)}\uparrow_1$ if it exists.
$\uparrow_{i}$ and $\downarrow_{i}$ respectively denote nearest neighbor upsampling and downsampling by average pooling for $2^i$ in each of the three spatial dimensions, except that for the binary mask $\vb*{m}$, the downsampling is minimum pooling to obtain binary values.

We also introduce a consistency loss to enhance the coarse-to-fine reconstruction process of the hierarchical network structure: for the output of each scale we downsample it and constrain its MSE loss against the target signal of the coarser scale.
The consistency loss is defined as
\begin{equation}
	\ell_{scale}^{(s)} =
	\left\|\vb*{m}\downarrow_{s}\odot\left(\text{clp}_\eta(f^{(s-1)}\downarrow_{1}) - \text{clp}_\eta(\vb*{u}\downarrow_{s})\right)\right\|^2, \quad s=1,2.
\end{equation}

\paragraph{Laplacian smoothness}
While going deeper backward the feature maps for different patches get shared more, the features close to output are by no means restricted for smoothness, which will in turn hamper the regularity of the final output shape.
To overcome this problem, we introduce a Laplacian smoothness loss on feature maps close to output: 
\begin{equation}
       \ell_{smooth}^{(s)} = \sum_{h\in P^{(s)}}\| \Delta \vb*{x}^{(h)}\|^2, \quad s=0,1,2
\end{equation}
where $\Delta$ is the Laplacian operator, applied on the feature maps of $P^{(s)}$ layers, which we choose to be the penultimate layer of the base block at scale $s$.

To summarize, we have the final objective function for training the completion network as 
\begin{equation}
    \ell_{total} = \sum_{s=0}^{2}{\frac{1}{\|\vb*{m}\downarrow_s\|_1}\left(\ell_{fit}^{(s)} + w_1 \ell_{smooth}^{(s)}\right)} + w_2 \sum_{s=1}^{2}{\frac{1}{\|\vb*{m}\downarrow_s\|_1}\ell_{scale}^{(s)}},
    \label{eq:total_loss}
\end{equation}
where the mask vector 1-norms $\|\vb*{m}\downarrow_s\|_1$ are used to normalize the squared losses of different scales, and $w_1=0.001, w_2=0.1$ are the weights for the two regularization terms chosen empirically. 
The fine scale output $f^{(0)}$ is taken as the final reconstructed shape.

\section{Implementation}
\label{sec:implementation}

In this section, we present more details about domain generation, network structures and the training process.

\paragraph{Completion domain generation}
To ensure the coverage of potential missing regions while also constraining the computational cost, we generate the sparse completion domain $\mathcal{M}$ (Sec.~\ref{sec:formulation}) in an incremental manner, summarized in Alg.~\ref{alg:mask_update}; the corresponding mask $\vb*{m}$ is computed by taking the difference $\mathcal{M}\setminus\Omega$.
For the input, we note that in addition to the narrow band sparse TSDF domain $\Omega$, there is a set of grid points $\mathcal{V}$ that are visible and known to be vacant during scanning.

\newlength{\textfloatsepsave} \setlength{\textfloatsepsave}{\textfloatsep} \setlength{\textfloatsep}{0pt} 
\begin{algorithm}[tb]
	\SetAlgoLined
	\SetKwFunction{Dilate}{Dilate}
	\SetKwFunction{FindGridsOnMeshBoundary}{FindGridsOnMeshBoundary}
	\SetKwFunction{GetLastPrediction}{GetLastPrediction}
	\SetKwFunction{GetAreaCloseToZero}{GetAreaCloseToZero}
	\SetKwFunction{ExcludeObserverdGrids}{ExcludeObserverdGrids}
	\SetKwInOut{Input}{Input}\SetKwInOut{Output}{Output}
	
	\Input{ $\Omega$, vacant set $\mathcal{V}$, number of dilation $N$(=4), $\eta$, \textit{UPDATE\_INTERVAL}(=250), \textit{MAX\_EPOCH}(=2k) }
	\Output{ [$\mathcal{M}_i$ for $i$-th epoch] }
	\For{$i\leftarrow 0$ \KwTo MAX\_EPOCH}{
		\uIf{$i==0$}{
			find grid points $\mathcal{B}$ bounding mesh open edges\;
			$\mathcal{M}_i$ = (\Dilate{$\Omega,N$}${\setminus}\mathcal{V}$) $\cup$ \Dilate{$\mathcal{B}$,2}\;
		}\uElseIf{$i{\%}$UPDATE\_INTERVAL $==0$}{
			get the latest predicted TSDF $\vb*{y}$\;
			find close-to-zero grids $\mathcal{Z} = \{(i,j,k) : |\vb*{y}_{i,j,k}| < \eta\}$\;
			$\mathcal{M}_i$ = \Dilate{$\mathcal{Z},N$}$\cup\Omega$\;
		}
		\Else{
			$\mathcal{M}_i=\mathcal{M}_{i-1}$;
		}
	}
	\caption{Sparse completion domain generation}
	\label{alg:mask_update}
\end{algorithm}
\setlength{\textfloatsep}{\textfloatsepsave}

As shown in Alg.~\ref{alg:mask_update}, the initial domain $\mathcal{M}_0$ is obtained by first dilating the known region $\Omega$ and then excluding the known vacant region $\mathcal{V}$, where dilation is the standard morphological operation \cite[Chapter~9]{gonzalez2002imageprocessing}. 
However, around the missing region boundaries that are tangential to scanning view directions, the visibility checking by excluding $\mathcal{V}$ might leave no grid points there to allow room for tangential completion.
Therefore, we also detect the grid points that bound the open edges of the incomplete mesh extracted from the input scan, and dilate them twice to obtain a padding layer to be included into $\mathcal{M}_0$. 
We have chosen the number of dilations to be 4 empirically, to balance between the sufficient coverage of the missing parts and the computational overhead induced by vacant voxels.

During the network optimization, we gradually update the mask based on the predicted surface.
In particular, after certain epochs of SGD optimization, we extract the latest predicted TSDF and detect the grid points that are close to the zero level set surface $\mathcal{Z}$ (Alg.~\ref{alg:mask_update}). 
By dilating the narrow band of $\mathcal{Z}$ as $\mathcal{M}$, we obtain the enlarged region on which there can be further potential surfaces.

After generating the completion domain for the finest scale as above, the domains for the two coarse levels (Fig.~\ref{fig:netstruct}) are built simply by maximum pooling, i.e., a coarse grid point is generated whenever one of its corresponding fine scale grid points exists.

\paragraph{Network details}
For cost-effective 3D computation, sparse convolution \cite{SparseConv2018} consumes sparse tensors defined on the generated completion domain, applies 3D convolution by filling zero for the unoccupied grid points, and outputs signals on the occupied sparse grid points only, which significantly saves GPU memory than naive 3D CNN and allows using TSDF of high resolution for modeling fine geometry details. 

Our three-scale network uses an encoder-decoder structure \textit{without} skip connections as the base block at each scale, as shown in Fig.~\ref{fig:netstruct}. 
By avoiding skip connections, the encoder feature maps are farther away from the output layers and induce more overlap of the deep features of different output patches, thus allowing their possible similarity. 
In addition, since the sparse convolution does not pass signals between grid points whose gaps are not bridged by the usually small convolution kernel size, the encoder by downsampling the sparse volume reduces the gaps in a coarse resolution and makes communication among disconnected points possible. 
Note that the detrimental effect of skip connections has also been noted in deep prior for image inpainting \cite{DeepImagePrior} albeit without further explanation.

\paragraph{Training with augmentation}
As discussed in Sec.~\mbox{\ref{sec:design}}, augmentation allows for more flexible shapes to be used for completing the missing regions.
We generate 23 augmented target TSDFs by rotating the original input randomly around the origin;
other augmentation operations may also be used by considering the specific object symmetries (see supplemental material for an example of mirror-reflection applied to bilaterally symmetric shapes), but we have used random rigid rotation for its pervasive applicability to diverse shapes.
The completion domains of these augmented shapes are generated individually according to Alg.~\ref{alg:mask_update}, and updated at a certain epoch simultaneously by their corresponding network predictions.
To speed up training, at each iteration (i.e. epoch in our case), we form a batch by randomly choosing 3 augmented shapes along with the original shape, and optimize the network to recover the 4 targets by one SGD step.
Empirically we find this is as effective as using all 23 augmentations in each epoch in terms of result quality, albeit with much lower computational cost.

\section{Results and discussion}
\label{sec:results}

\begin{figure}
	\centering
	\includegraphics[width=\linewidth]{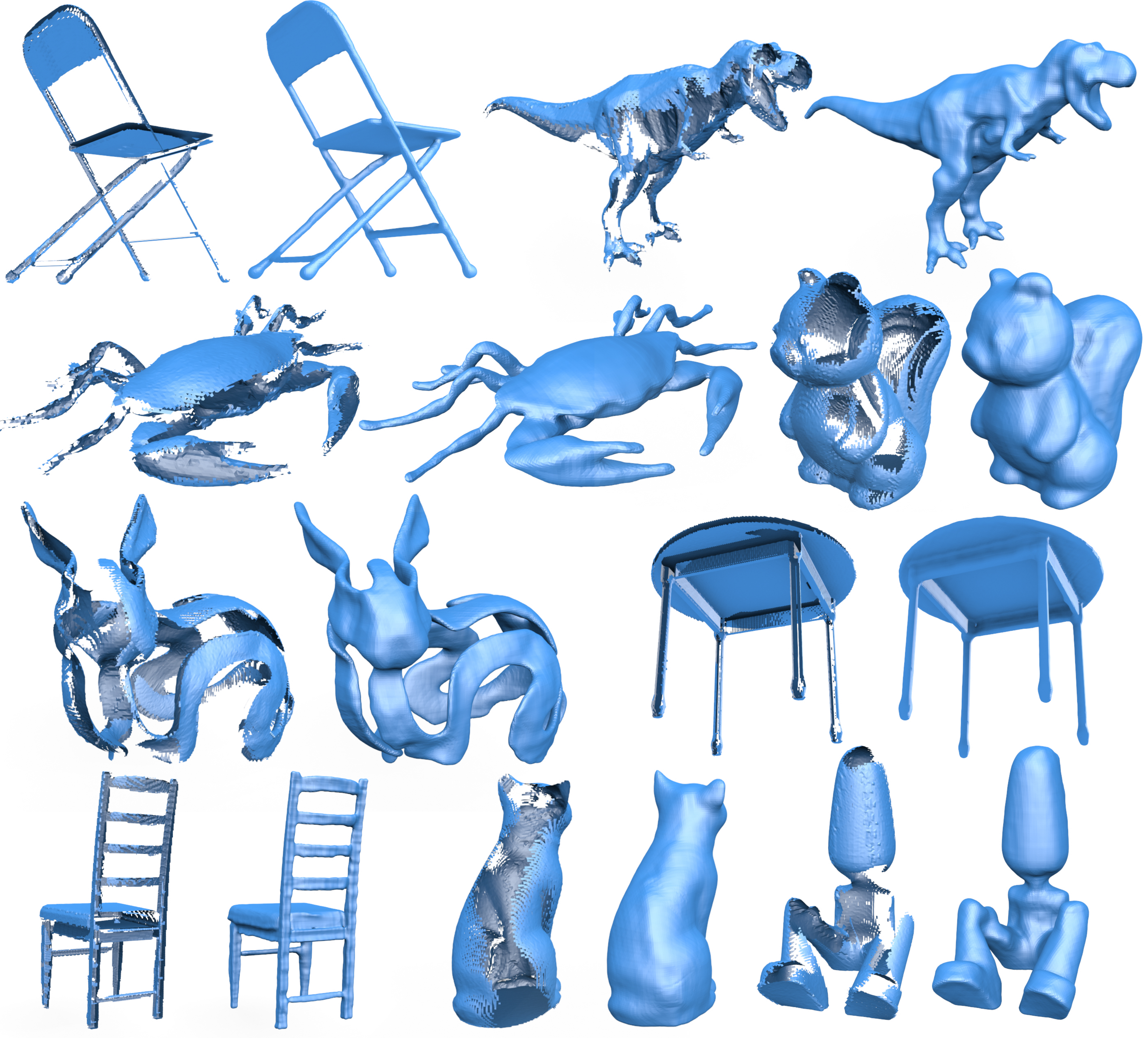}
	\vspace{-6mm}
	\caption{More results of shape completion by our unsupervised method. The diverse objects have been collected from ShapeNet, Thingi10k and Free3D, with challenging features like slim bars, thin layers and large missing regions. Our method plausibly recovers the features by automatically exploiting the self-similarities and overall coherence with known parts.}
	\label{fig:gallery}
	\vspace{-3mm}
\end{figure}

By using our unsupervised completion networks, we have recovered plausible shapes from input partial scans of diverse categories.
Some of the visual results are shown in Figs.~\ref{fig:teaser}, \ref{fig:recon_process}, \ref{fig:gallery}, \ref{fig:ablation}, \ref{fig:comparison}.
Next we present the details of experiments, extensive ablation tests for validating the framework design, and comparisons with previous methods.

\subsection{Setup}
\label{sec:setup}

\paragraph{Test data generation.} 
In the experiments, we randomly picked diverse categories of testing models from existing datasets, including chairs and lamps from ShapeNet Core v2 \cite{ShapeNet15}, free-form shapes from Stanford 3D Scanning Repository \cite{VolumeIntegration_SIG96}, 3D printing models from Thingi10k \cite{Thingi10K} and models from Free3d \cite{free3d}. 
Following \cite{Berger2013,3depn,HanComplete2017}, we synthesize the partial scans by simulating scanning through rendering depth maps in $3{\sim}4$ camera viewpoints that are placed randomly but evenly on the bounding sphere of a model by furthest sampling, 
and fusing the maps with volumetric TSDF integration \cite{VolumeIntegration_SIG96} implemented in Open3D \cite{open3d}.
In particular, each depth map is first converted to a TSDF whose value for each voxel on a view ray is computed as its distance along the ray to the observed pixel depth value, subject to truncation; the TSDFs for all depth maps are then integrated by weighted averaging. Meanwhile, the voxels observed to be empty under any viewpoint are included in $\mathcal{V}$ (Sec.~\ref{sec:implementation}).
The TSDFs are discretized with resolution $256^3$.

For the ablation and comparison tests, we have randomly selected a set of 30 representative test models, with 4 chairs, 3 lamps, 3 freeform shapes, and benches, airplanes, monitors, missiles and taps, each of 4 models. In addition, although our method is fully unsupervised, we have chosen hyperparameters like iteration number empirically based on the first 10 models as the validation set, and the rest 20 models are used only for test.
Examples are shown in Figs.~\ref{fig:ablation} and \ref{fig:comparison}; the complete set and more results are provided in the supplemental material.

\paragraph{Evaluation metric}
Following \cite{hanocka2020point2mesh,what3d_cvpr19,knapitsch2017tanks}, we evaluate the numerical quality of results by F-score, which computes the harmonic mean of precision and recall between the predicted and ground truth point clouds, with point pairs under distance threshold $\delta$ considered correct matching.
We set $\delta{=}0.7\%$ of the volume side length, which is tight enough for distinguishing good reconstructions from rough approximations.

\paragraph{Runtime}
We use the SparseConv framework \cite{SparseConv2018} implemented in PyTorch for building our networks and training.
The networks are optimized for 2000 epochs with the Adam solver \cite{kingma2014adam} at a fixed learning rate $2\times 10^{-3}$. 
On a machine with Nvidia GeForce GTX 1080 Ti GPU and Intel Xeon Silver 4108 CPU of 1.8GHz, the optimization process takes about 4 hours for completing a model.

\subsection{Ablation tests}
\label{sec:ablation}

\begin{figure*}
	\centering
	\begin{overpic}[width=\linewidth]{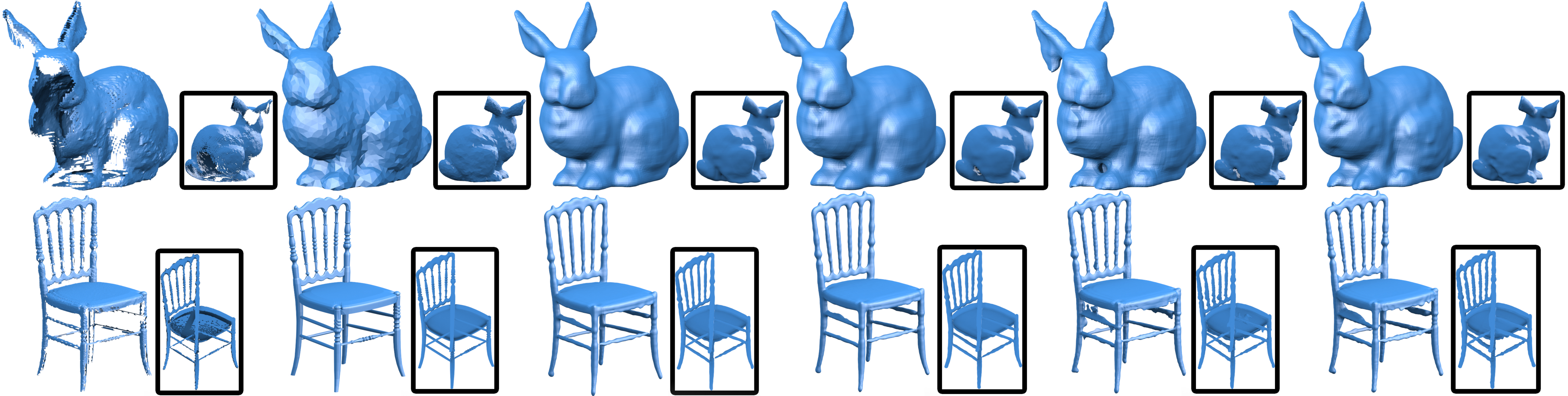}{
			\put(2,-1.5){\small (a) input partial scan}
			\put(19.5,-1.5){\small (b) ground truth}
			\put(36,-1.5){\small (c) full model}
			\put(52.5,-1.5){\small (d) w/o hierarchy}
			\put(69,-1.5){\small (e)  w/o Laplacian}
			\put(86,-1.5){\small (f) w/o augmentation}
	}\end{overpic}
	\vspace{-3mm}
	\caption{Ablation test results on two examples. (c) the full model result is close to the ground truth despite the large missing regions. (d) without multi-scale hierarchy, holes appear at the bunny tail due to difficulty for bridging the large gaps there. (e) without Laplacian smoothness, local regularity is lost and extraneous parts emerge. (f) without augmentation, the completed shapes are less fit and plausible.}
	\label{fig:ablation}
	\vspace{-1mm}
\end{figure*}
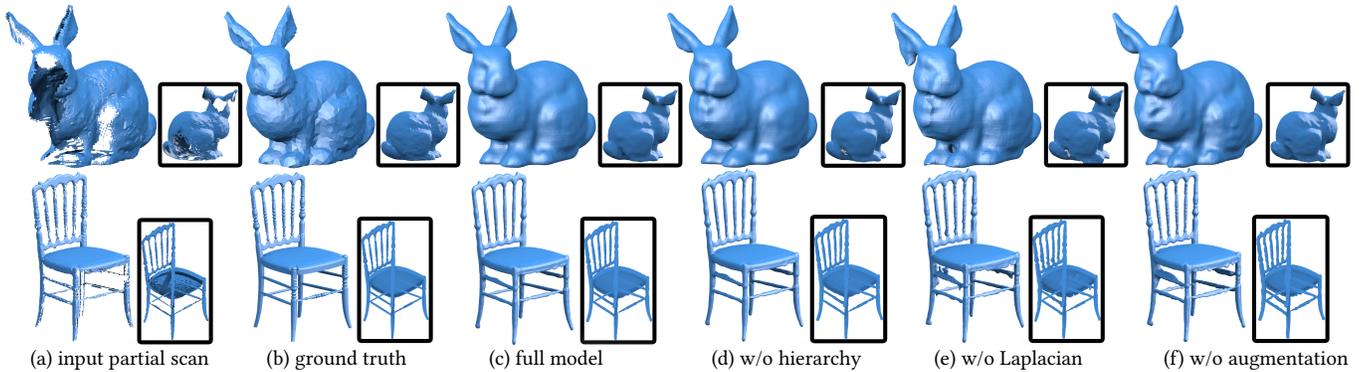

In this section, we do ablation tests on various factors of the 3D CNN and its training, to validate the design choices made in Sec.~\ref{sec:design} as well as the corresponding implications of NTK through experiments.

\paragraph{Multi-scale hierarchy}
As discussed in Sec.~\ref{sec:design}, the hierarchical network structure is adopted to generate the geometrical structures in a coarse-to-fine manner, thus capturing both large scale structures and fine level details. 
In contrast, by using a single fine scale network for completion, as shown in Fig.~\ref{fig:ablation}(d), holes appear around the bunny tail and a beam is missing on the chair. 
Quantitative evaluation in Tab. \ref{tab:fscore_ablation} also shows that the accuracy decreases when using a single scale network.

\paragraph{Network depth}

It is predicted by NTK that the success of the deep prior on shape completion relies on the correlation of neighborhood patches.
In Sec.~\ref{sec:ntk_interp} we have shown that deeper networks would lead to larger NTK correlation through more shared deep features, and thus more coherent and generally better shape reconstructions. 
To validate if this is true, we remove the factor of multi-scale hierarchy and focus on testing the base block network structure of the fine scale (Fig.~\ref{fig:netstruct}). 
We compare it against a shallow but widened model, which has approximately the same amount of trainable parameters as the single scale baseline, for the sake of fair comparison. 
Note that the base block has also been trimmed in terms of intermediate feature channel sizes, so that the shallow but wide network can fit into the limited GPU memory.
In particular, the trimmed base block network has an encoder of form $[e_{0,j}]_{j=1,\cdots,5}$, with each $e_{0,j}$ of feature size 16, and the decoder mirror reflected, while the shallow network has an encoder of $[e_{0,j}]_{j=1,\cdots,2}$ with corresponding feature sizes [16, 32] (more details about convolution kernel sizes in Appendix \ref{appdx:network_architecture});
as a result, the shallow network has slightly more trainable parameters than the deep network.
However, the reconstruction accuracy given in Tab. \ref{tab:ablation_depth} shows that decreasing the network depth leads to significant drop on the completion performance.

\begin{table}
	\centering
	\begin{tabular}{ccccc}
		\hline
		Configuration&ave&std&min&max\\
		\hline
		Baseline & $\mathbf{96.36}$ & $\mathbf{3.27}$ & $\mathbf{88.66}$ & $99.95$\\
		w/o hierarchy & $95.40$ & $4.06$ & $85.09$ & $\mathbf{99.98}$\\
		w/o Laplacian & $91.25$ & $5.27$ & $78.90$ & $99.54$\\
		w/o augment & $93.05$ & $5.28$ & $83.26$ & $99.70$\\
		\hline
	\end{tabular}
	\caption{F-score of different ablation configurations tested on 30 models.}
	\vspace{-7mm}
	\label{tab:fscore_ablation}
\end{table}

\begin{table}
	\centering
	\begin{tabular}{ccccc}
		\hline
		&ave& std& min & max\\
		\hline
		Deep network & $\mathbf{94.68}$ & $\mathbf{4.80}$ & $\mathbf{80.22}$ & $\mathbf{99.82}$\\
		Shallow network & $92.73$ & $6.94$ & $71.34$ & $99.82$ \\
		\hline
	\end{tabular}
	\caption{Network depth impacts reconstruction accuracy. The deep and shallow models are single scale networks with similar amounts of trainable parameters. But the deep model has better accuracy.}
	\vspace{-7mm}
	\label{tab:ablation_depth}
\end{table}

\paragraph{Laplacian smoothness}

As discussed in Sec.~\ref{sec:design}, the Laplacian smoothness is applied to penultimate layers of feature maps to improve the local regularity of the reconstructed surface. 
This is helpful especially because closer to the output layers the overlapping of different patches is smaller, and so is their coherence.
As shown in Fig.~\ref{fig:ablation}(e), removing the smoothness term would lead to extraneous surfaces on missing regions, for example around the bunny's ear and the chair beam. 
A decrease of F-score accuracy is also observed in Tab.~\ref{tab:fscore_ablation} without the smoothness term.

\paragraph{Early stopping}

\begin{figure}[tb]
	\centering
	\vspace{4mm}
	\begin{overpic}[width=0.7\linewidth]{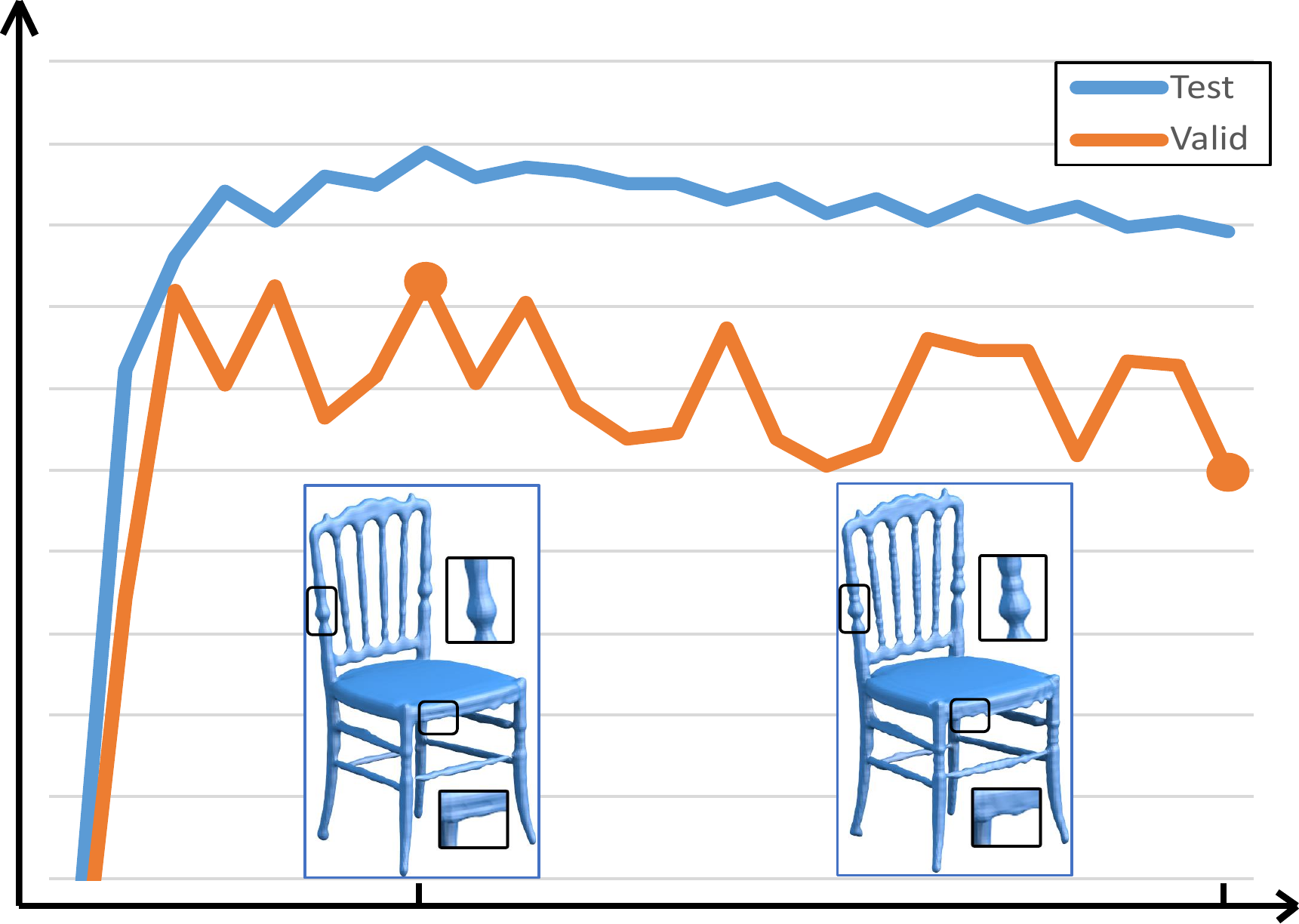}{
			\put(-6.3, 2){\small 88}
			\put(-6.3, 13.5){\small 90}
			\put(-6.3, 26){\small 92}
			\put(-6.3, 39){\small 94}
			\put(-6.3, 51){\small 96}
			\put(-6.3, 63.5){\small 98}
			\put(-6.3, 71.5){F-score$(\%)$}
			\put(28,-4){\small 2k}
			\put(91,-4){\small 6k}
			\put(96,3.5){\#epoch}
	}\end{overpic}
	\vspace{-1mm}
	\caption{The average F-score with respect to the number of training epochs, evaluated on 10 validation models and 20 test models, respectively. The shapes quickly recover during the early epochs, reach stable around 2k, but then slowly degrade by fitting to the extra details with more epochs. This demonstrates the necessity of early stopping.
	Results for an example chair model in the validation set at the two marked iteration counts are shown in the inset; the input scan and ground truth shape are shown in Fig.~\ref{fig:ablation}.}
	\label{fig:early_stop}
	\vspace{-3mm}
\end{figure}
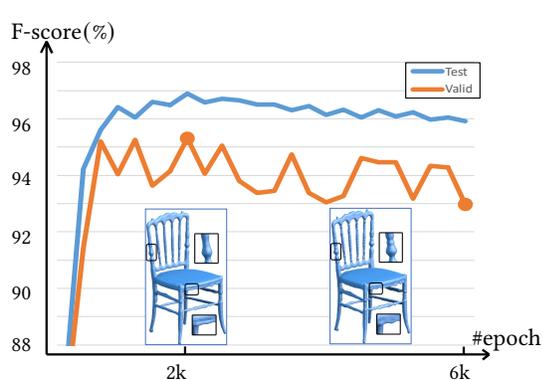

The NTK interpretation of deep prior implies that early stopping would help the network capture dominant features but avoid focusing on extra details, to improve the regularity of the predicted shape \cite{NTK_NIPS2018}.
Intuitively, the kernel gradient descent (Eq.~\ref{eq:kernel_gradient}) induced by network training would adjust the network predictions fastest along the dominant principal eigenvectors of the kernel matrix, which correspond to the major structural similarities among the shape patches; after that, further SGD iterations would encourage superfluous self-similarities instead, which is usually undesirable.
Fig.~\ref{fig:early_stop} shows how the average F-score tested on 10 validation models changes with respect to the training epochs. 
It can be seen that the curve first increases rapidly, indicating the fitting of the dominant shape features by the networks. 
But then the curve decreases slowly after 2k epochs, due to the loss of regularity resulting from the fitting of extra details. 
A similar trend is observed for the 20 test models, demonstrating the appropriateness of the chosen  hyperparameter value.
The reconstructed shapes of a chair model at 2k and 6k epochs are also shown in the figure, from which we see how the 6k epoch result on one hand captures more subtle details like the blobs on back beams, but on the other hand undesirably replicates such bulges to the other regions of completion.
Note that similar phenomenon is also empirically observed by deep prior for image reconstruction \cite{DeepImagePrior}.

\paragraph{Data augmentation}

\begin{figure}
	\vspace{4mm}
	\centering
	\begin{subfigure}[b]{0.23\textwidth}
		\centering
		\begin{overpic}[width=\textwidth]{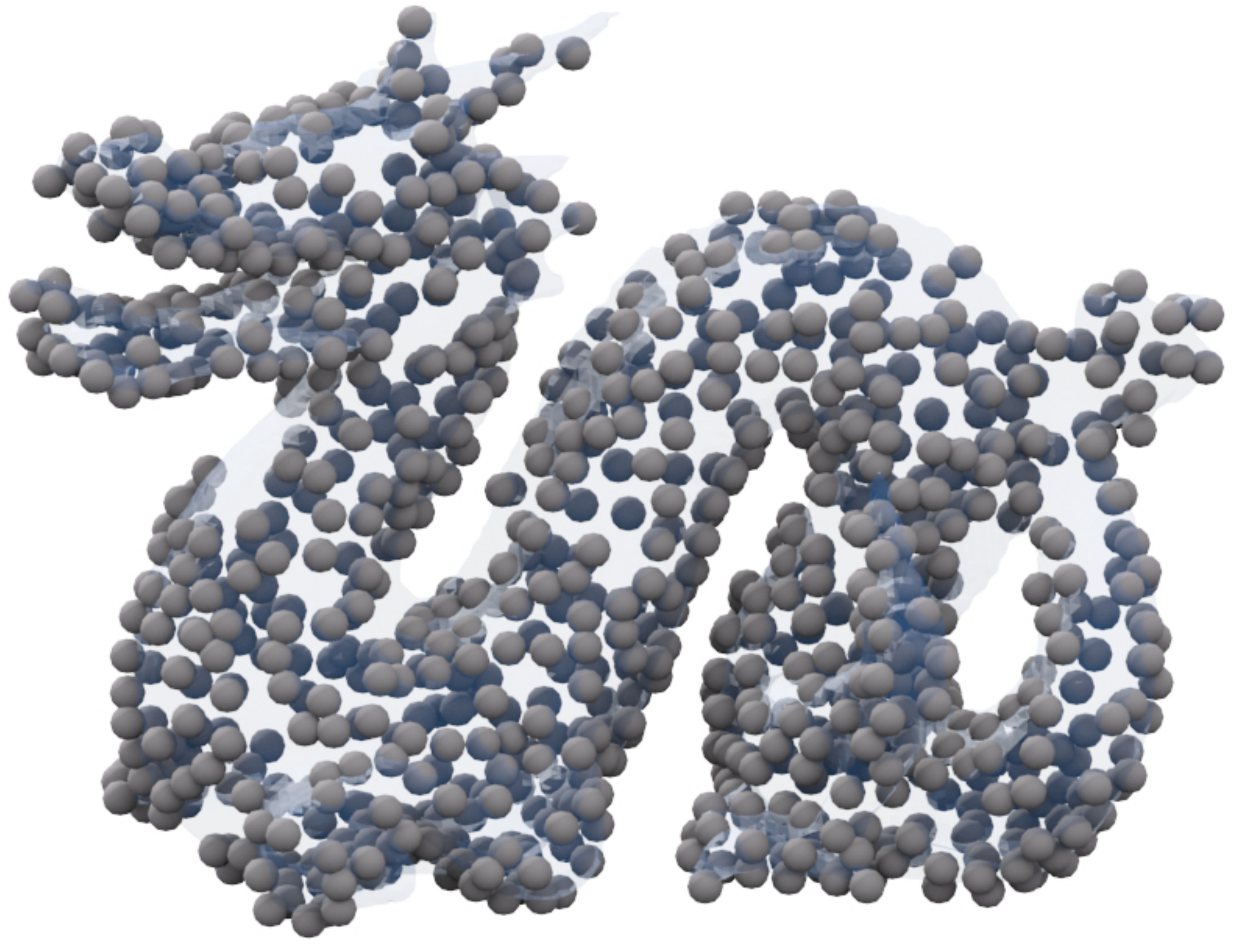}
			\put(63,62){\frame{\includegraphics[width=0.35\textwidth]{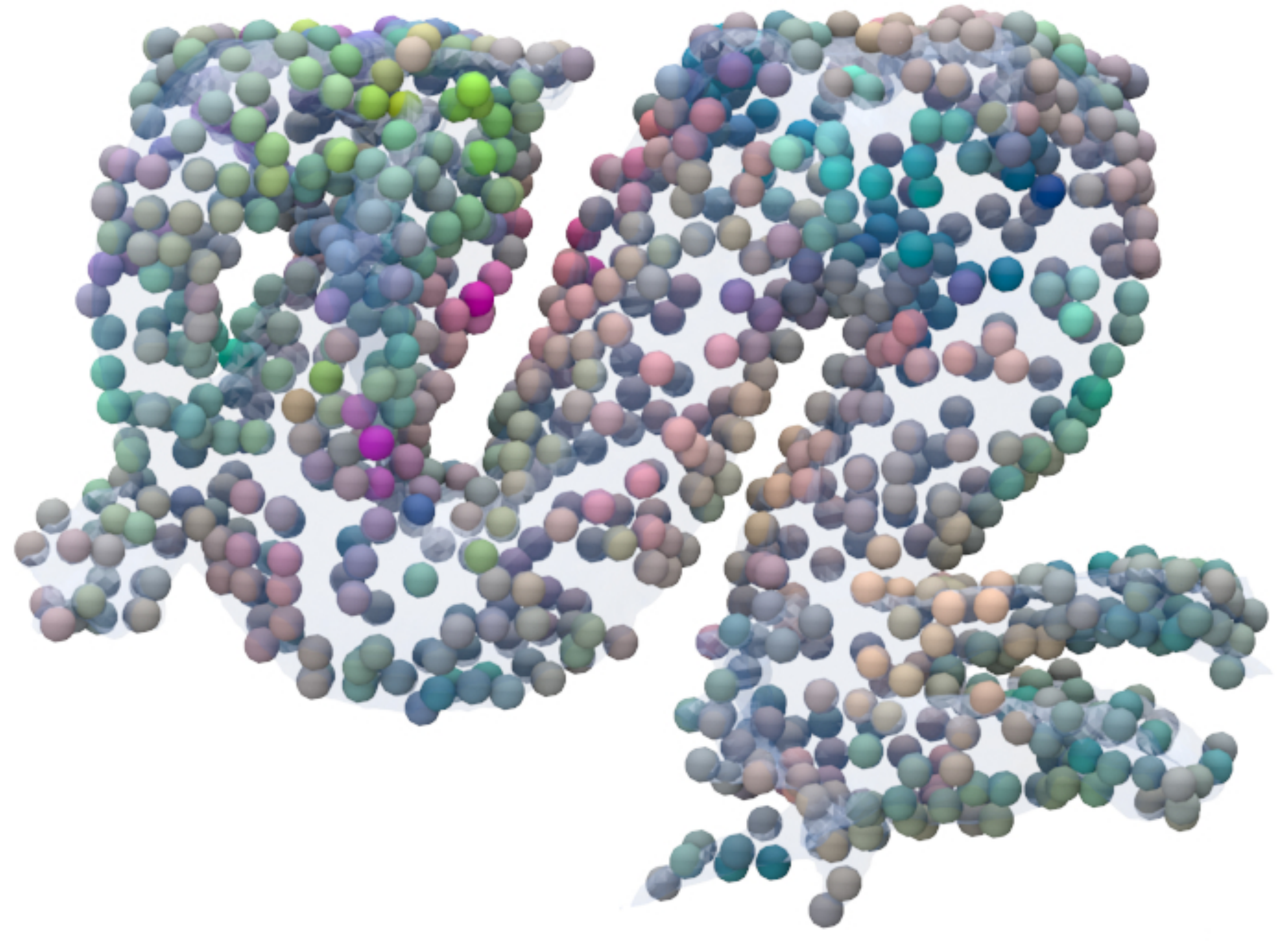}}}
		\end{overpic}
		\vspace{-5mm}
		\caption{}
		\label{fig:NTK_wtAug_vis}
	\end{subfigure}
	\hfill
	\begin{subfigure}[b]{0.23\textwidth}
		\centering
		\begin{overpic}[width=\textwidth]{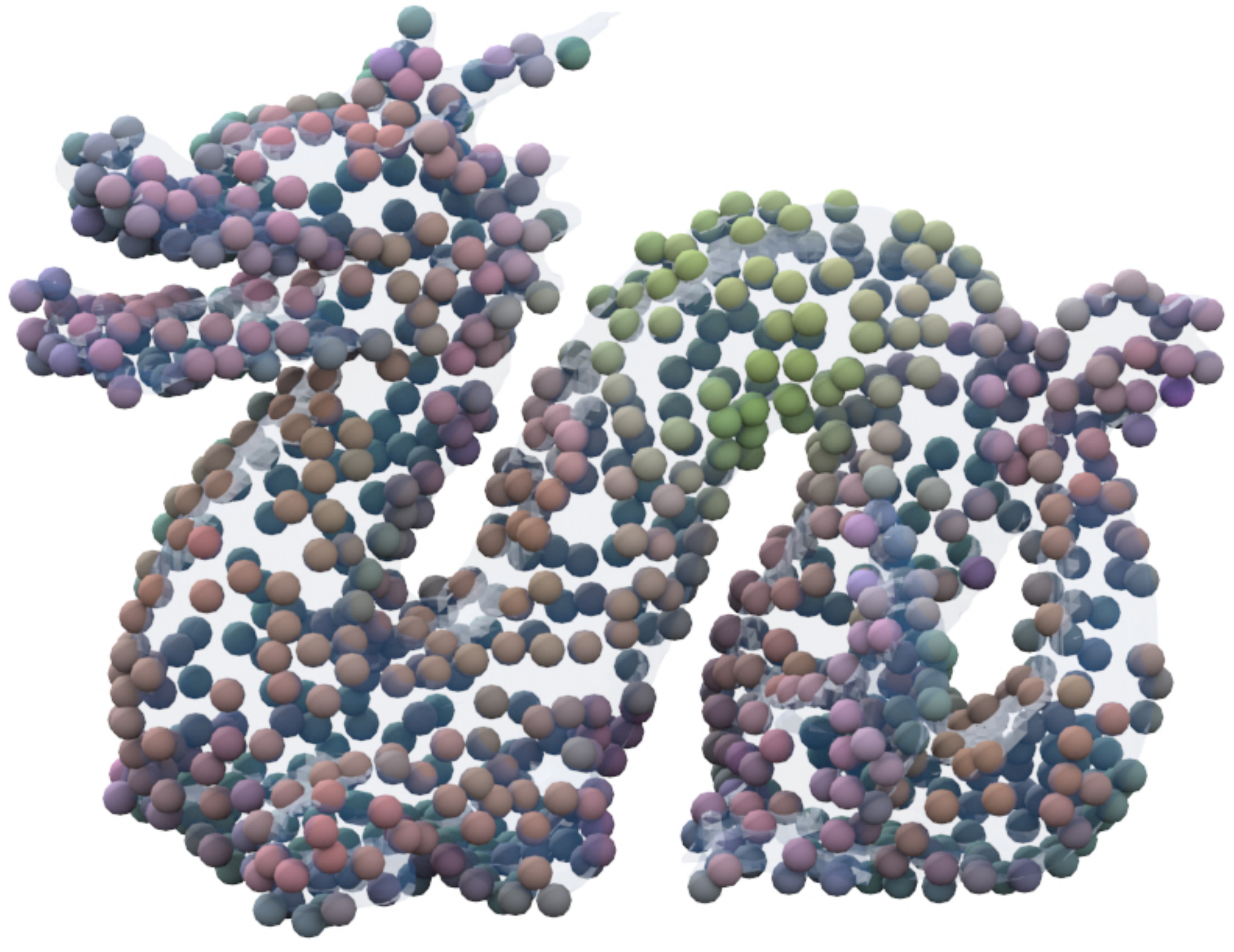}
			\put(63,62){\frame{\includegraphics[width=0.35\textwidth]{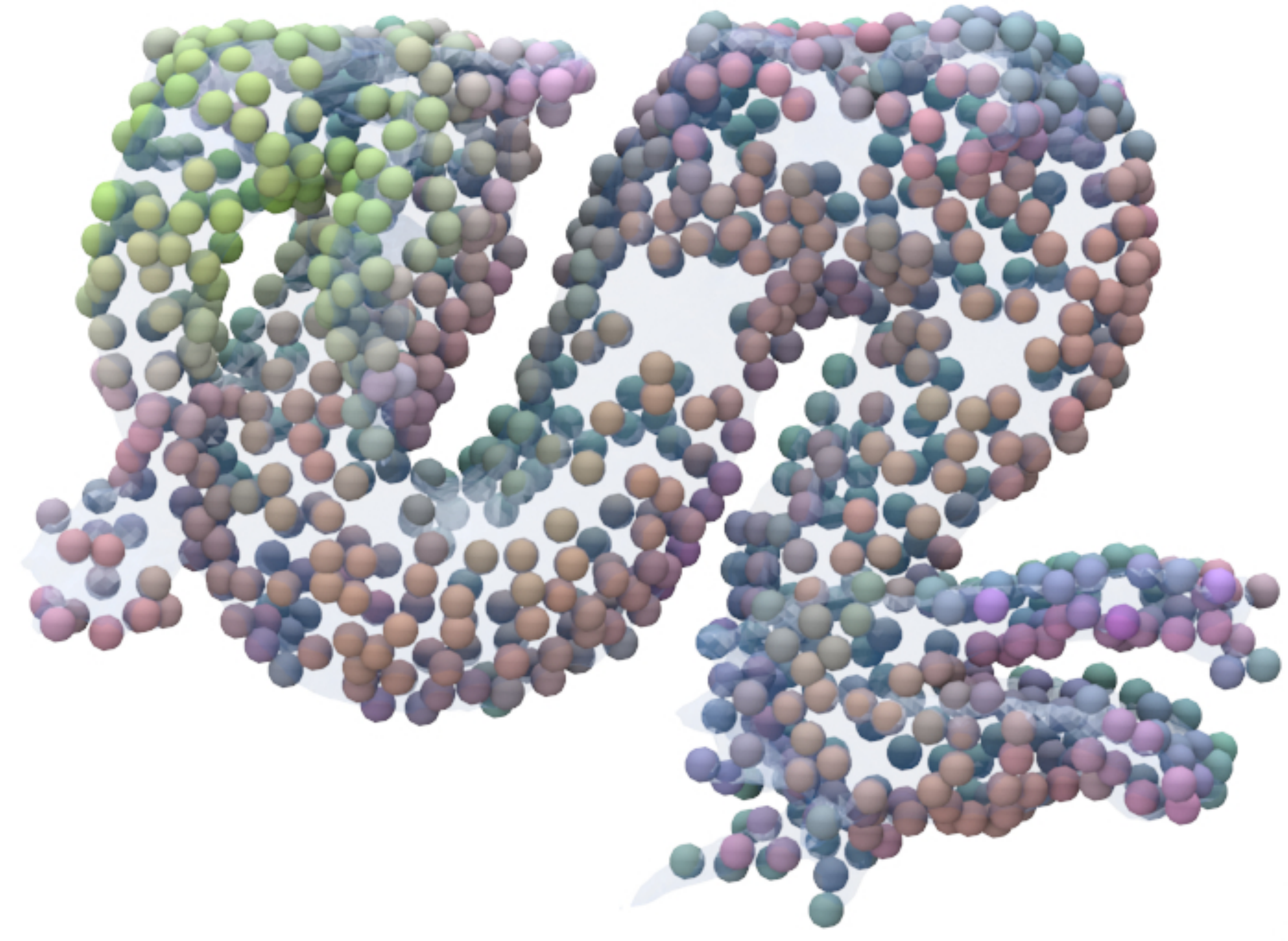}}}
		\end{overpic}
		\vspace{-5mm}
		\caption{}
		\label{fig:NTK_noAug_vis}
	\end{subfigure}
	\hfill
	\begin{subfigure}[b]{0.23\textwidth}
		\centering
		\includegraphics[width=\textwidth]{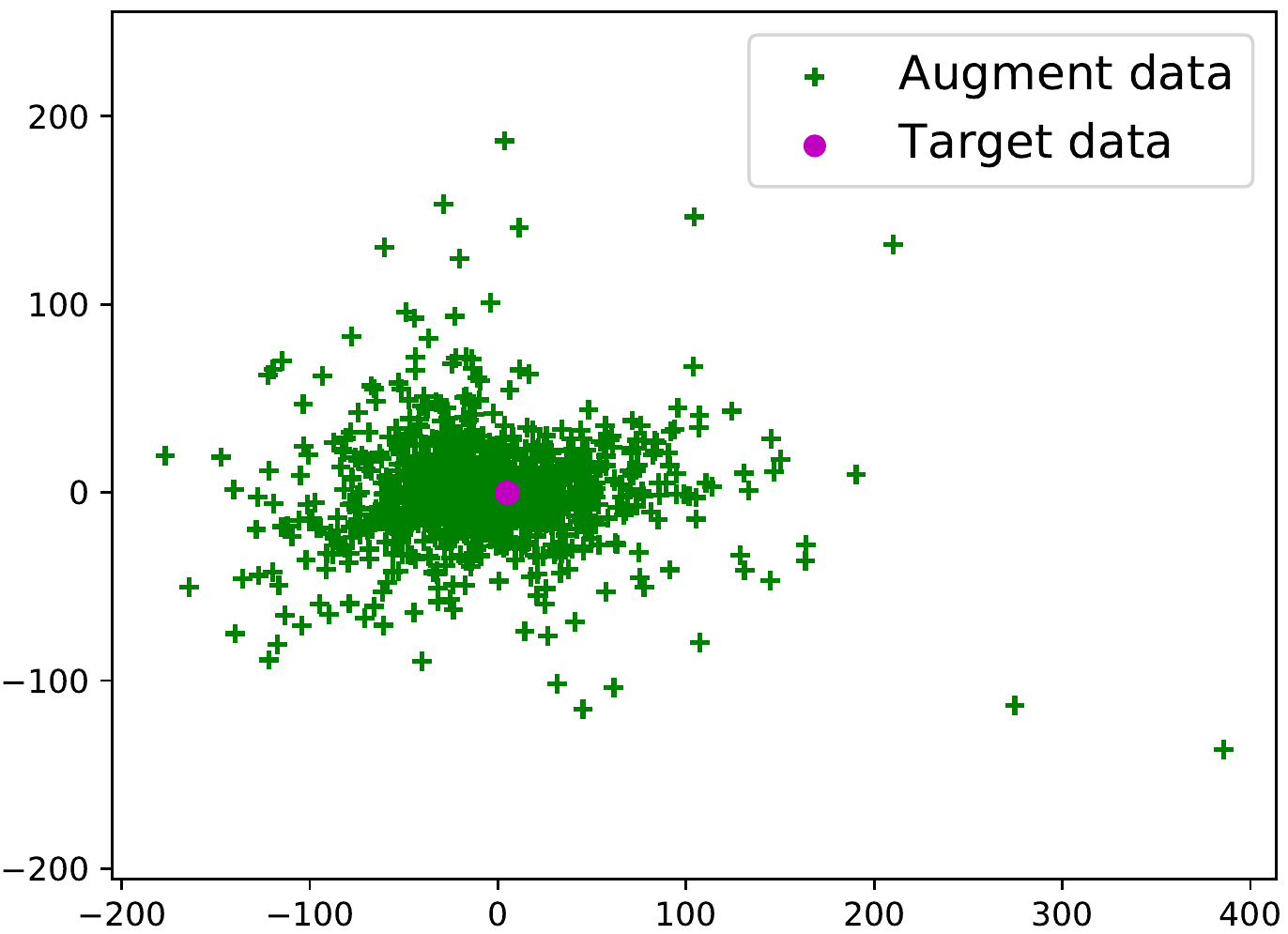}
		\vspace{-5mm}
		\caption{}
		\label{fig:NTK_wtAug_scatter}
	\end{subfigure}
	\hfill
	\begin{subfigure}[b]{0.224\textwidth}
		\centering
		\includegraphics[width=\textwidth]{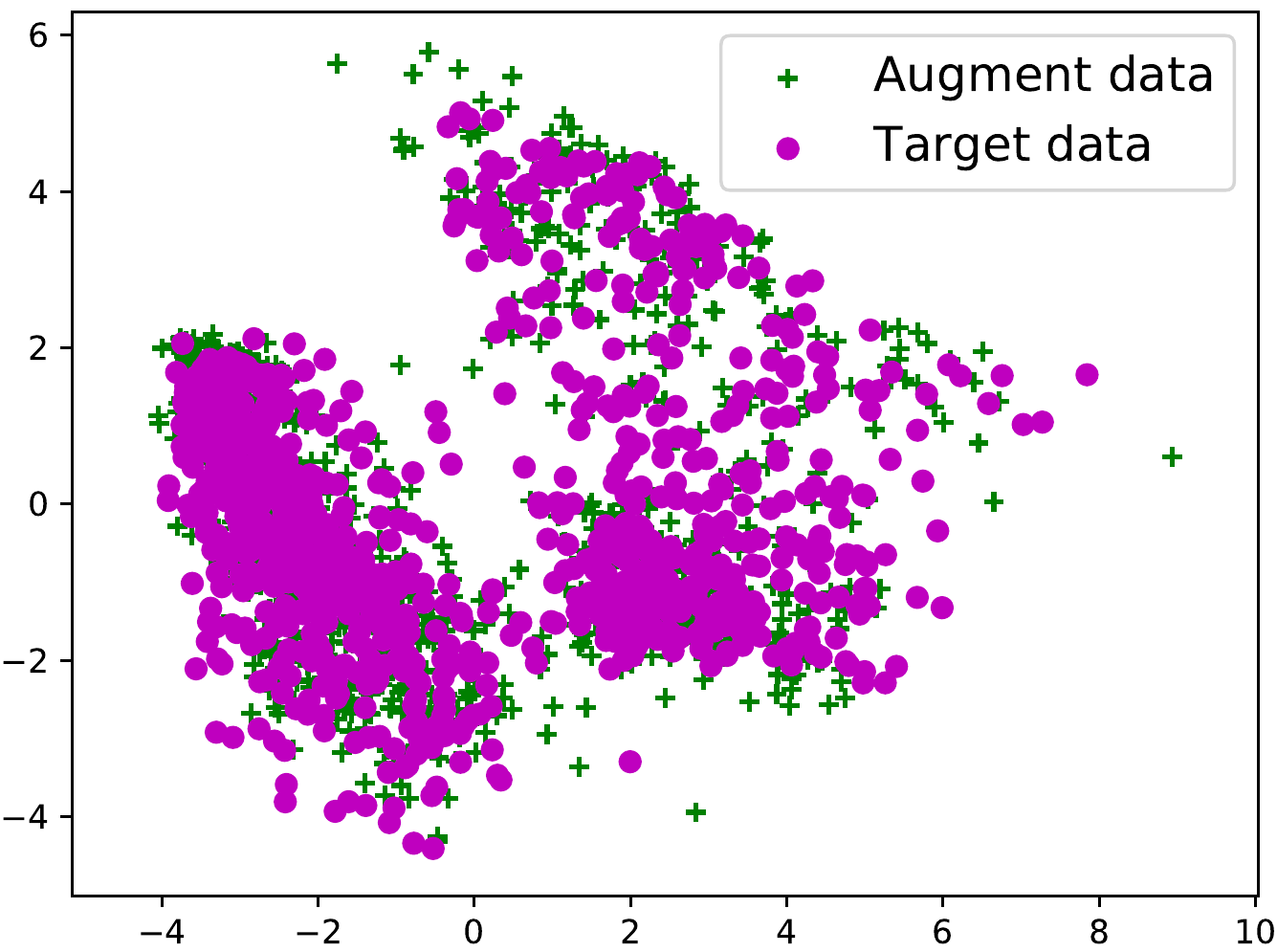}
		\vspace{-5mm}
		\caption{}
		\label{fig:NTK_noAug_scatter}
	\end{subfigure}
	\vspace{-3mm}
	\caption{Kernel PCA analysis with NTK for 2k sample points, half from the target models (a),(b) and half from the augmented models (upper right corners of (a),(b)), trained without (left) and with augmentation (right). 
	The scatter plot and color coding are in the same format as Fig.~\ref{fig:NTK_vis_before_after_opt}.
	Without augmentation, the two point sets show no correlation under NTK; with augmentation, the two sets have strong correlations.}
	\label{fig:NTK_aug}
	\vspace{-3mm}
\end{figure}

The augmentation of training data by randomly rotating the input shape expands the NTK feature space with more training samples (Eq.~\ref{eq:kernel_regression}), thus making it easier to fill the missing region with more flexible and plausible patches. 
To validate its efficacy, we disable the augmentation and obtain results of lower accuracy, as shown in Tab.~\ref{tab:fscore_ablation}. 
The correlation of input shape and augmented copies can also be visually inspected directly.
As shown in Fig.~\ref{fig:NTK_aug}, we apply kernel PCA dimension reduction to the 2k sample points from the target shape and the augmented shape with their NTK features.
When the network is trained solely on the input shape, the augmented points are entirely unrelated to the target shape points in NTK distance (Fig.~\ref{fig:NTK_aug}(c)) and exhibit a much larger range random distribution than the concentrated target points; 
this is also shown in the color coding by the top three principal coordinates in Fig.~\ref{fig:NTK_aug}(a), where the target points show a single color, as compared with the diverse but irregular colors of the augmented points.
However, when the network is trained with augmentations, the points from both the augmented and the target shapes are highly correlated, with overlapped 2D distributions (Fig.~\ref{fig:NTK_aug}(d)), and color codings that show how similar structures at different parts and across the models are related (Fig.~\ref{fig:NTK_aug}(b), see again the similar points at two horns and the green body parts at different locations of the target and augmented models).

\subsection{Comparison}
\label{sec:comparison}

\begin{figure*}[t]
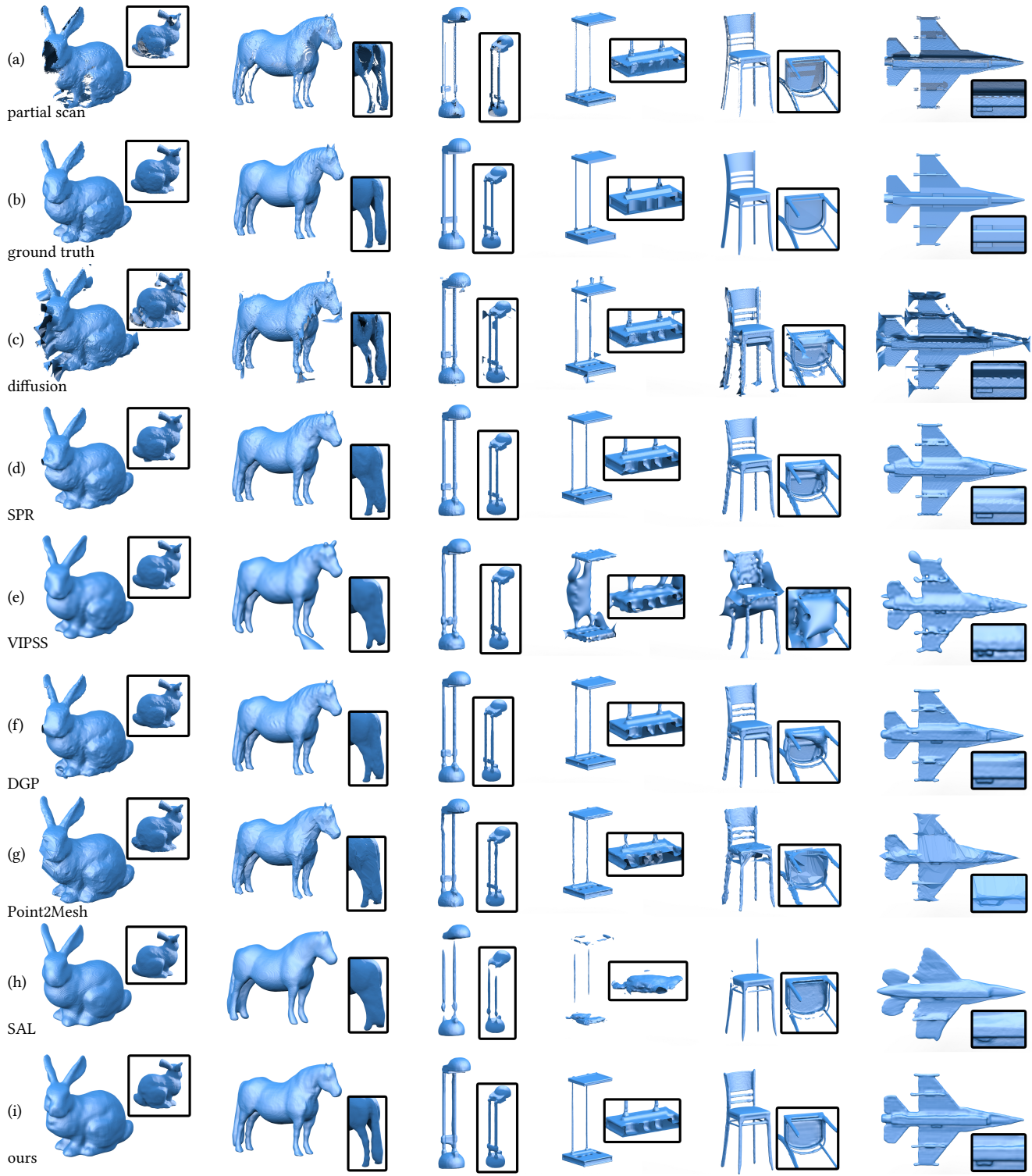

	\centering
	\begin{overpic}[width=0.95\linewidth]{comp}{
			\put(-2,95){\small (a)}	\put(-2,90.5){\small partial scan}
			\put(-2,83){\small (b)} \put(-2,78.5){\small ground truth}
			\put(-2,71){\small (c)} \put(-2,67){\small diffusion}
			\put(-2,60){\small (d)} \put(-2,56){\small SPR}
			\put(-2,49){\small (e)} \put(-2,45){\small VIPSS}
			\put(-2,38){\small (f)} \put(-2,33){\small DGP}
			\put(-2,27){\small (g)} \put(-2,22){\small Point2Mesh}
			\put(-2,16){\small (h)} \put(-2,12){\small SAL}
			\put(-2,5){\small (i)}  \put(-2,1){\small ours}
	}\end{overpic}
	\vspace{-2mm}
	\caption{Comparison with category agnostic methods that apply smoothness prior on traditional spatial models ((c) diffusion, (d) SPR, (e) VIPSS), or rely on inherent regularities of deep neural networks ((f) DGP, (g) Point2Mesh, (h) SAL).}
	\label{fig:comparison}
	\vspace{-3mm}
\end{figure*}

We compare our approach with representative reconstruction methods of three types.
The first type contains the traditional methods that do not rely on deep learning but directly optimize the shapes for smoothness, in particular the volumetric diffusion \cite{DavisDiffusion2002}, screened Poisson reconstruction (SPR) \cite{kazhdan2013screened} and variational implicit point set surface (VIPSS) \cite{VIPSS2019}.
The second type is the methods that use inherent regularities of neural networks, including deep geometric prior (DGP) \cite{Williams_2019_CVPR} and Point2Mesh \cite{hanocka2020point2mesh}.
The third type is the deep learning based methods supervised by labeled categorical data sets, for example the state-of-the-art DeepSDF \cite{DeepSDF}. 
Quantitative evaluations and visual comparisons among the methods are shown in Tab.~\ref{tab:fscore_compare} and Fig.~\ref{fig:comparison}.

\begin{table}[tb]
	\centering
	\begin{tabular}{ccccc}
		\hline
		Method&ave&std&min&max\\
		\hline
		Diffusion& $80.66$ & $5.60$ & $73.40$ & $95.12$ \\
		SPR & $94.42$ & $5.30$ & $83.05$ & $99.91$ \\
		VIPSS & $65.32$ & $32.50$ & $8.63$ & $99.72$ \\
		DGP$^*$ & $87.61$ & $12.84$ & $49.25$ & $99.82$ \\
		Point2Mesh & $83.03$ & $15.70$ & $43.83$ & $99.53$ \\
		DeepSDF$^{**}$ & $75.83$ & $9.49$ & $64.27$ & $85.51$ \\
		SAL & $78.93$ & $21.76$ & $27.52$ & $98.74$ \\ 
		Our& $\mathbf{96.36}$ & $\mathbf{3.27}$ & $\mathbf{88.66}$ & $\mathbf{99.95}$ \\
		\hline
	\end{tabular}
	\caption{F-score of different methods tested on 30 models. $^*$ DGP is tested on 25 models, with 5 left due to out of GPU memory. $^{**}$ DeepSDF is tested on the 4 chair models of the test set, as it uses category specific training. }
	\vspace{-7mm}
	\label{tab:fscore_compare}
\end{table}

\vspace{-1mm}
\paragraph{With volumetric diffusion \cite{DavisDiffusion2002} and screened Poisson reconstruction \cite{kazhdan2013screened}} Volumetric diffusion and screened Poisson reconstruction are two classical non-learning based methods for hole filling and surface reconstruction, both solving elliptic PDEs on the 3D volume that lead to smooth implicit functions whose zero level set surfaces are the reconstructed surfaces.
As such, they can produce smooth and regular results for relatively small missing areas. 
However, for large missing areas, the surface smoothness prior may be insufficient, so that they may compute surfaces that do not respect larger structural regularities or are even unclosed, as shown in Fig.~\ref{fig:comparison}(c)(d) and the high discrepancies from ground truth shapes in Table~\ref{tab:fscore_compare}.

\vspace{-1mm}
\paragraph{With VIPSS \cite{VIPSS2019}} 
VIPSS is a most recent implicit function based surface reconstruction method.
While conceptually similar to screened Poisson reconstruction, VIPSS uses a different smoothness energy and models the implicit function by RBF based kernel interpolation discretized on a sparse set of input points.
The combination leads to the benefit of convex optimization and robustness even with very few sample points, but also comes with the cost of directly solving large dense systems that become intractable for more sample points.
In the comparisons, we have sampled evenly 1k points for VIPSS from the input partial scans.
As can be seen through Fig.~\ref{fig:comparison}(e) and Table~\ref{tab:fscore_compare}, VIPSS shares similar problems as the other direct geometric optimization methods, failing to produce closed shapes for larger missing regions and  unaware of larger scale structural regularities, as dictated by the surface smoothness prior.

\vspace{-1mm}
\paragraph{With Deep Geometric Prior \cite{Williams_2019_CVPR}} 
Deep Geometric Prior models a shape as a collection of local patches, with each patch mapped from a canonical chart parameterization through its specific MLP network, and overlapping patches bearing compatible transitions across the charts. 
The per-patch regularity is therefore due to the intrinsic MLP smoothness. 
DGP reconstructs a shape from input point cloud by fitting the local patches to partitioned regions of the point cloud, guaranteeing cross-patch compatibility along the way.
The final surface model is then extracted by sampling the patches densely followed by screened Poisson reconstruction.
By applying DGP on the point clouds of partial scans (Fig.~
\ref{fig:comparison}(f)), however, we find that the approach on one hand generates smooth and tight fitting shapes to the known point clouds, but on the other hand is unaware of the structural similarities of different parts and cannot generate reasonable completions for large missing areas.
Numerical results tested on 25 of the 30 models (Table~\ref{tab:fscore_compare}) also confirm its lower accuracy compared with our method; the 5 models excluded run out of GPU memory for DGP.

\vspace{-1mm}
\paragraph{With Point2Mesh \cite{hanocka2020point2mesh}} 
Point2Mesh is a recent method that makes use of the regularity prior of a mesh CNN \cite{hanocka2019meshcnn} to reconstruct meshed model from point clouds. 
The method deforms an initial mesh containing the input point cloud to ``shrink-wrap'' over the points, refining and deforming the mesh by CNN adaptively along the way.
It is observed that the shared convolution weights over different mesh patches lead to structural similarities especially for geometric textures.
While such a shrink-wrap process may be sufficient for reconstructing high quality meshes from unevenly sampled point clouds, it is however not very suitable for completing larger missing regions.

For one thing, since the surface topology is unchanged throughout the process, a wrong initial guess due to too much missing data would inevitably lead to poor results, as shown in the chair example of Fig.~\ref{fig:comparison}(g). 
In comparison, our TSDF shape representation is flexible in changing topology.
In addition, the edge-based convolution of mesh CNN is a 2D manifold-based operation dependent on mesh tessellation.
On the other hand, over the missing regions there is no guidance about tessellation, and the structural similarity with known regions are more about the 3D embedding than mere offset from a base 2D manifold. 
Therefore, we hypothesize that while the edge-based convolution induces localized self-similarity for geometric textures over surface patches, it is not suitable for shape completion.
Indeed, as shown in Fig.~\ref{fig:comparison}(g), for complex shapes (e.g. chair and lamp) or shapes with large missing area (e.g. bunny and horse), though the result meshes of Point2Mesh are guaranteed to be watertight, the features on the restored area may not be desirable, especially for thin and concave regions, because few guiding points are there to drive the shrink-wrap process, and the self-similarity of local surface patches seems to be unaware of larger scale structural regularities to induce plausible completions.
Numerical accuracy shown in Table~\ref{tab:fscore_compare} is also generally worse than our results.
However, it may be possible that Point2Mesh can build on our results and reconstruct a mesh representation that has sharper geometric textures, which we leave as future work to investigate.

\begin{figure}
	\centering
	\begin{overpic}[width=\linewidth]{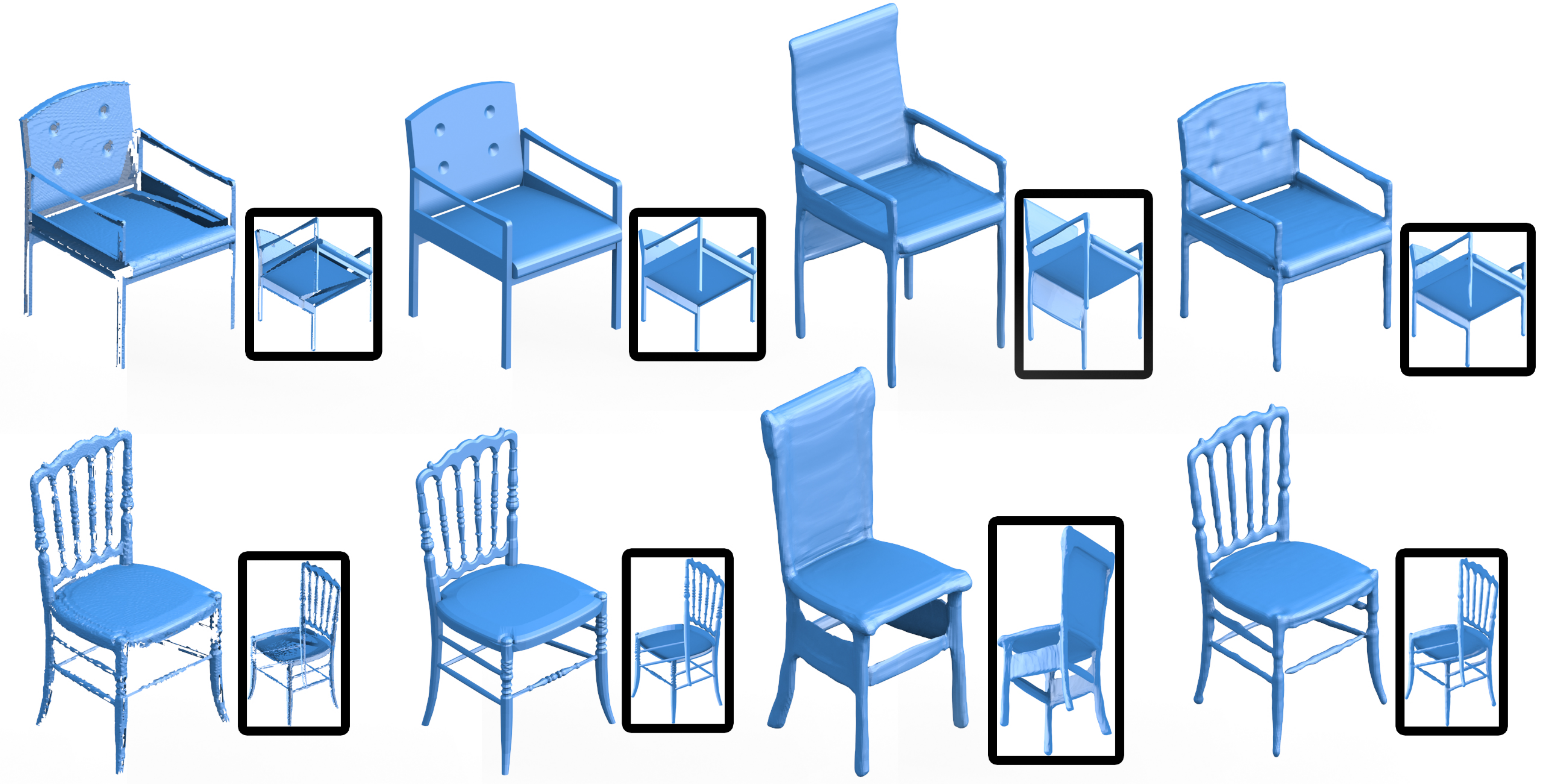}{
			\put(3,-2.5){\small (a) partial scan}
			\put(28,-2.5){\small (b) ground truth}
			\put(53,-2.5){\small (c) DeepSDF}
			\put(79,-2.5){\small (d) ours}
	}\end{overpic}
	\vspace{-4mm}
	\caption{Comparison with the supervised DeepSDF \cite{DeepSDF} that is trained on the chair category of ShapeNet. While DeepSDF fails to reconstruct close matches for the input shapes unseen during training, our unsupervised approach recovers them faithfully. }
	\label{fig:comparison_dsdf}
	\vspace{-5mm}
\end{figure}
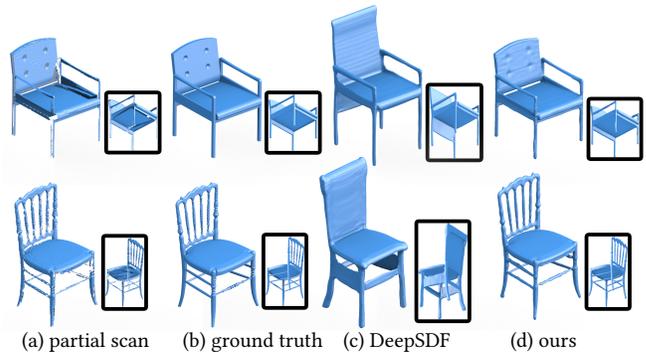

\vspace{-1mm}
\paragraph{With DeepSDF \cite{DeepSDF}} 
DeepSDF is a supervised learning-based method that learns to encode each shape in a category as an implicit SDF function parameterized by a tuple: the deep MLP network, the object instance code and the spatial coordinate where the implicit function is evaluated.
While DeepSDF can autoencode a large category of objects such that each object is recovered with high fidelity, 
its learned category-specific prior may not sufficiently adapt to input partial scans in the 3D completion setting, especially for unseen data during training.
This is shown in Fig.~\ref{fig:comparison_dsdf}: given the partial scans with important structures, DeepSDF however fails to preserve the structures and instead reconstructs chairs which despite completeness resemble little of the input scans.
In contrast, our unsupervised approach can complete the missing parts while being fully adaptive to the known regions.

\vspace{-1mm}
\paragraph{With sign agnostic learning \cite{atzmon2020sal}}
Sign agnostic learning (SAL) is a latest implicit function based 3D reconstruction method. 
Like DeepSDF, it parameterizes the SDF function with an MLP network, but proposes to use sign agnostic objective functions to enable reconstruction from point sets without normal vectors. 
In addition, similar to us, SAL focuses on fitting to specific shapes without category prior by optimizing the network from scratch.
However, despite the inherent smoothness of MLP generated shapes, it cannot capture contextual cues for shape completion.
To compare with SAL by experiments, we use the dense point cloud of the partial scans as its input.
As shown in Fig.~\ref{fig:comparison}(h), throughout the diverse reconstructions, SAL produces smooth shapes but fails to capture the thin structures of lamps, chair and aircraft, probably due to the sparse points around those regions, and fails to generate important details for the completed regions in the bunny, horse and aircraft models, which is likely due to its lack of context modeling. 
The over-smoothness is also translated into larger numerical errors, as shown in Table~\ref{tab:fscore_compare}.
In comparison, our method naturally captures the multi-scale contexts and generates completions that are similar in pattern to the known shapes.

\textbf{Remark.} We note that from the NTK perspective, in order to enhance an MLP for contextual awareness and shape completion, both the input spatial coordinates and the intermediate latent features should likely be transformed into a multi-scale representation, whose components change at different rates spatially, such as the Fourier coefficients, so that the NTK feature embedding (Eq.~\ref{eq:kernel_function}) can have improved consistency across scales. 
Indeed, \cite{FourierFeatures2020} analyzes how replacing the input features with Fourier coefficients enhances high frequency reconstruction, while \cite{Siren2020} shows that periodic activation functions help detail generation as well. 
An analysis of spectrum modulation focused on shape completion would be interesting to explore in the future.


\subsection{Limitations}
In this section, we explore how the proposed method works under different conditions to know more about its limits.
In particular, we empirically study the impacts of symmetry, missing data ratio, volume resolution and input noise level.

\begin{figure}[t]
	\centering
	\begin{overpic}[width=\linewidth]{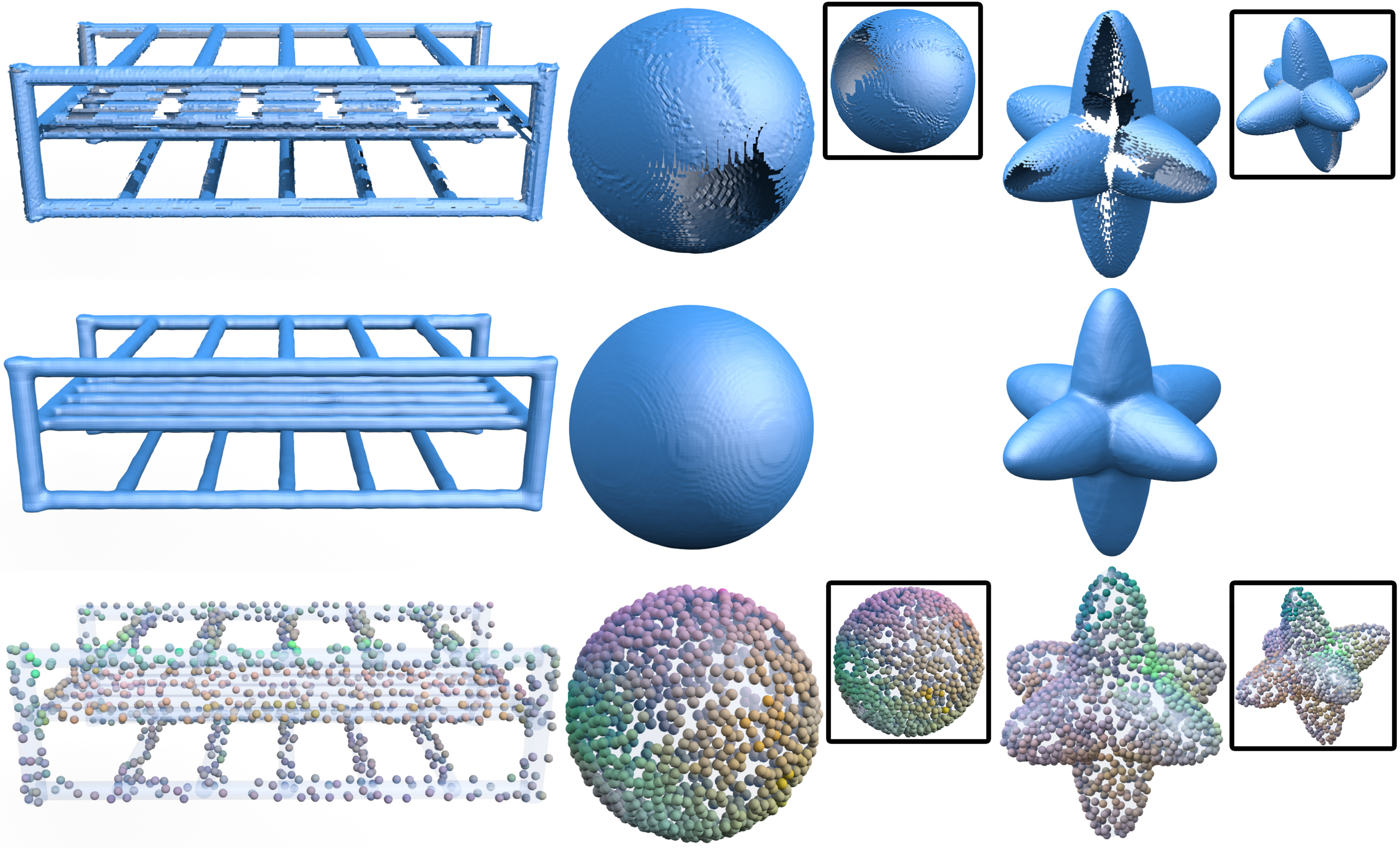}{
			\put(16,-2.5){\small (a)}	
			\put(48,-2.5){\small (b)} 
			\put(84,-2.5){\small (c)}
	}\end{overpic}
	\vspace{-4.5mm}
	\caption{Completing symmetric shapes. (a) translational symmetry, (b) and (c) rotational symmetry. The insets show augmentations by random rotation. From the kernel PCA color coding in the last row, it can be seen that translational symmetries along and across the poles are captured by the CNN, while rotational symmetry is achieved by correspondences between the target and the rotation-augmented models.}
	\label{fig:symm}
	\vspace{-4.5mm}
\end{figure}

\begin{figure}[t]
	\centering
	\begin{overpic}[width=0.75\linewidth]{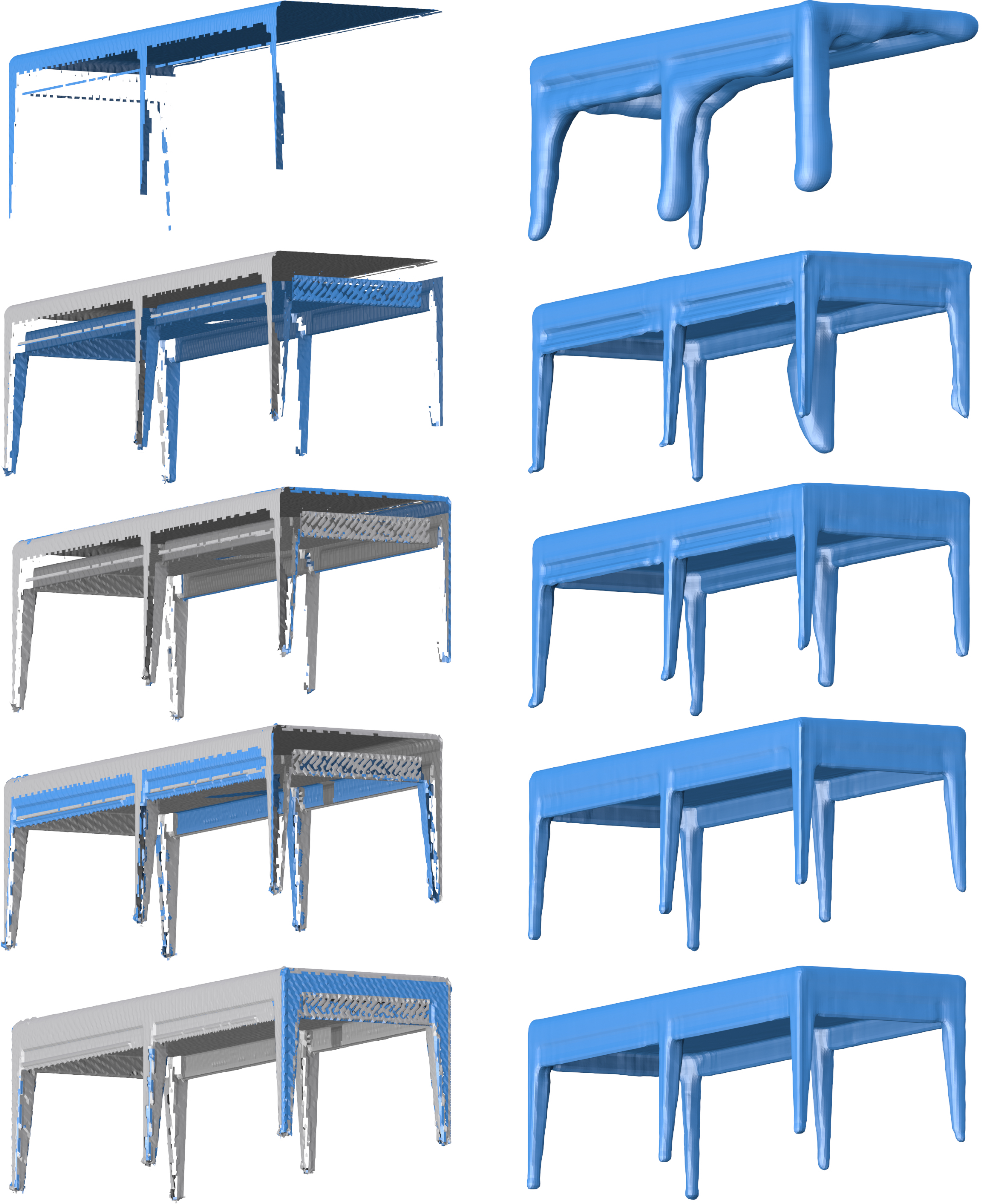}{
			\put(-13,85){\small \#cam=1}
			\put(-13,65){\small \#cam=2}
			\put(-13,45){\small \#cam=3}
			\put(-13,25){\small \#cam=4}
			\put(-13,5){\small \#cam=5}
	}\end{overpic}
	\vspace{-3mm}
	\caption{Different ratios of missing data. From top to bottom, the number of observed depth images is increased from 1 to 5. For each partial scan in the left column, the blue part is the newly added observation and the gray part denotes the previous observations. Reconstructions on the right column show steady improvement with the increasing data.}
	\label{fig:missing_degree}
	\vspace{-3mm}
\end{figure}

\begin{figure}
	\centering
	\begin{overpic}[width=0.9\linewidth]{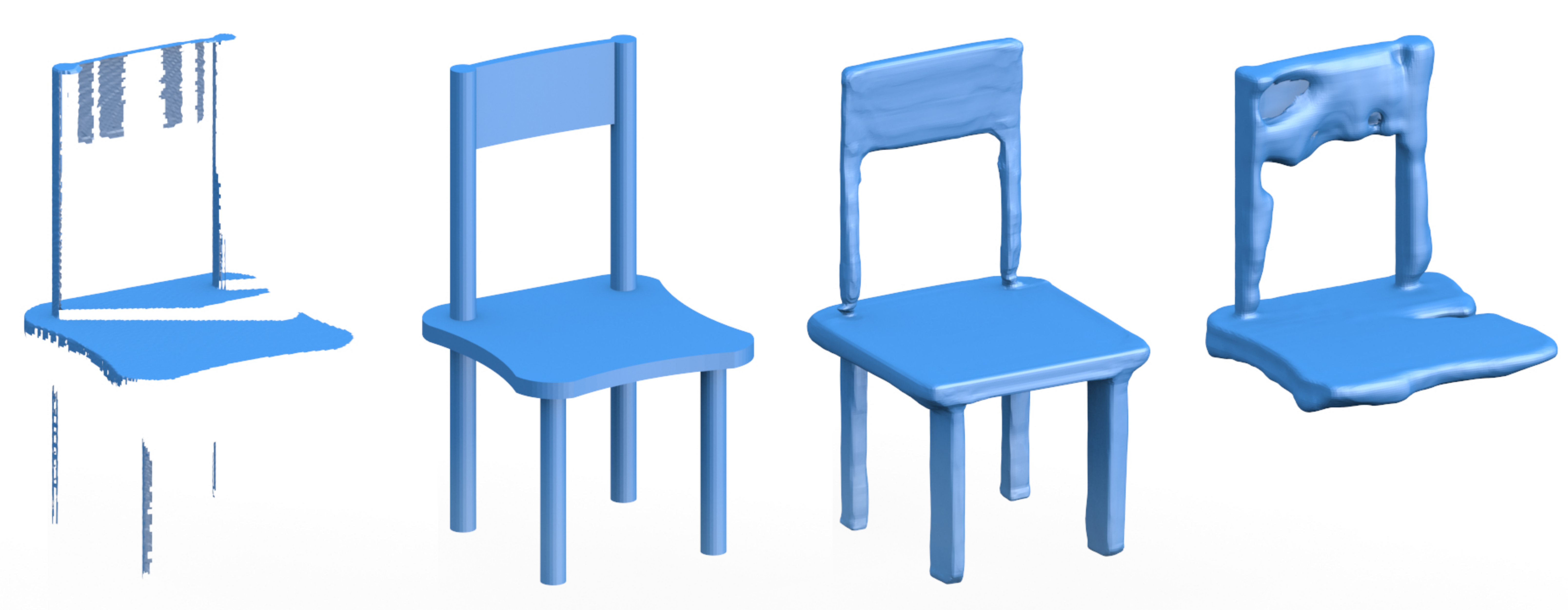}{
			\put(0,-2){\small (a) partial scan}
			\put(25,-2){\small (b) ground truth}
			\put(53,-2){\small (c) DeepSDF}
			\put(79,-2){\small (d) ours}
	}\end{overpic}
	\vspace{-1mm}
	\caption{Handling extremely incomplete scan. While DeepSDF can utilize the strong categorical prior to reconstruct a complete chair given the few input, our method without dataset prior cannot work well in this case.}
	\label{fig:failure}
	\vspace{-3mm}
\end{figure}
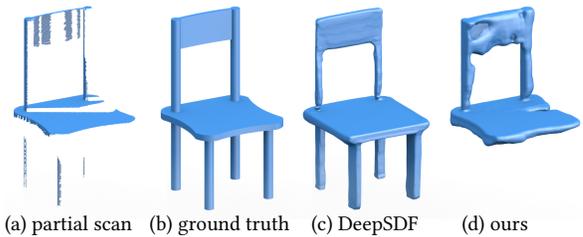

\paragraph{Symmetry} Unlike previous methods that explicitly detect symmetry and enforce it in the reconstruction \cite{pauly2008discovering,Mitra2013}, our method implicitly encodes repetitive patterns across the scales. For the examples with translational and rotational symmetries in Fig.~\ref{fig:symm}, our results are shown to preserve the symmetries well, although no explicit guarantees can be made.
In particular, by inspecting the kernel distances of sampled points, we can see that the 3D CNN detects translational symmetries along the poles and across the poles that reside on same layers, but the rotational symmetry is achieved through similarities with the rotated augmentations; this aligns with the fundamental translation equivariance property of CNNs \cite{lecun98}. 
In addition, we note that even disconnected regions (Fig.~\ref{fig:symm}(a), poles in the middle layer) can be connected by the completion.
However, if a whole component of the symmetry group is missing without any hints in the input, our method may not recover it; see the missed table legs in Fig.~\ref{fig:missing_degree} and the following discussion.

\paragraph{The ratio of missing data} 
As an unsupervised and class-agnostic method, our approach naturally degrades as the ratio of missing data increases. 
An example is shown in Fig.~\ref{fig:missing_degree}, where as more observations are made, the reconstruction gets more complete.
In particular, we note that without any hints around two table legs (top row, only one observation), the completion misses the entire components, as our self-similarity based reasoning is unaware of semantics or explicit symmetry rules. 
This is in contrast to methods that learn categorical shape priors by supervised training, which always search in the feasible shape space and produce a complete shape even with little conditioning.
As shown in Fig.~\ref{fig:failure}, for such a limited partial input, our approach does not give meaningful result, but DeepSDF recovers a proper chair nonetheless. 
Therefore, we see our unsupervised category-agnostic approach as a complement to the supervised methods, and consider it an important future work to combine both approaches for achieving robustness to little data and adaptiveness for known data simultaneously.

\begin{figure}[t]
	\centering
	\begin{overpic}[width=\linewidth]{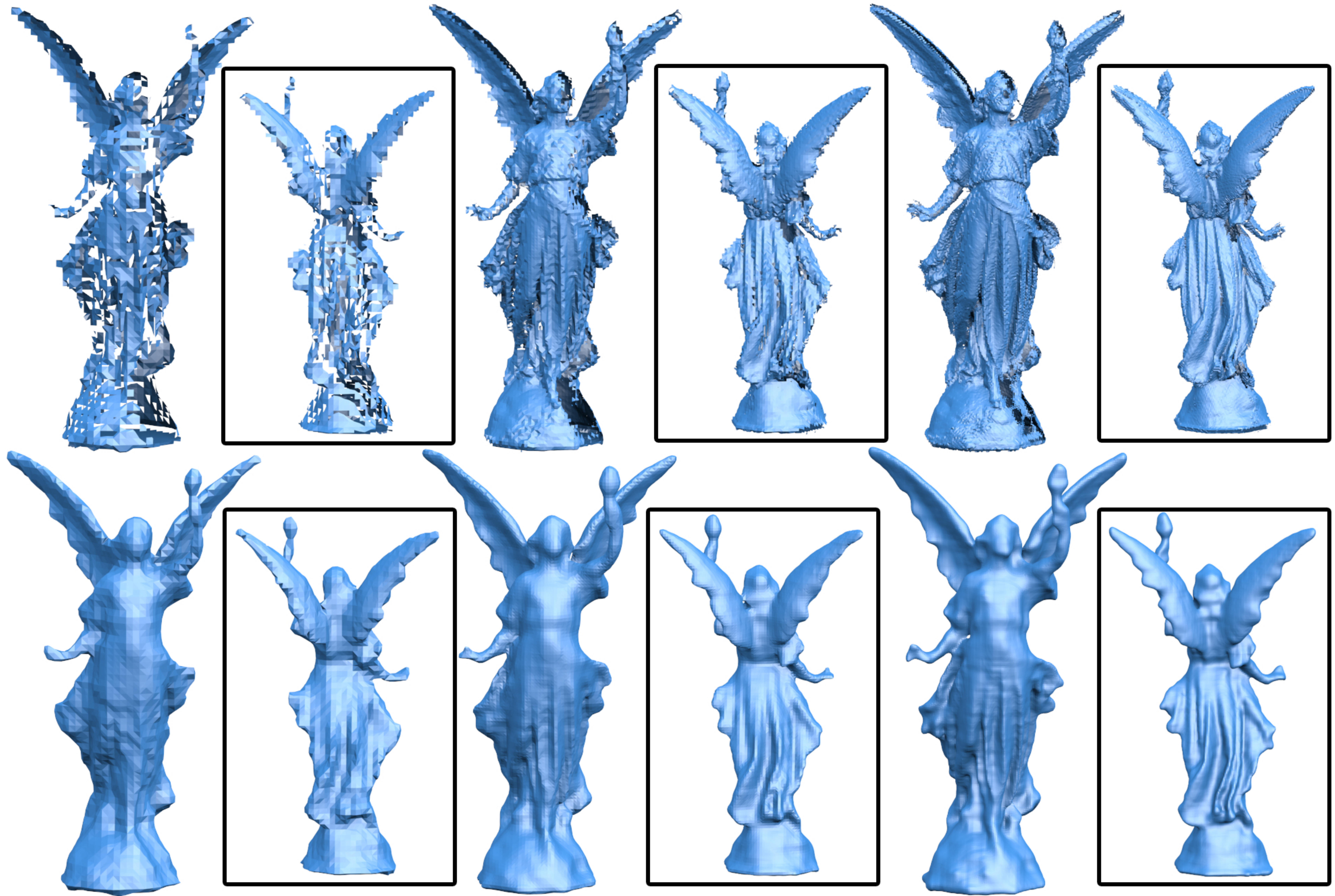}{
			\put(11,-4){\small (a) $128^3$}
			\put(43,-4){\small (b) $256^3$}
			\put(75,-4){\small (c) $512^3$}
	}\end{overpic}
	\vspace{-3mm}
	\caption{Reconstruction with different resolutions of volume grid. The same set of depth images are integrated into TSDF volumes of specified resolutions, on which the completion network is applied by scaling its per-layer spatial resolution correspondingly. Note how the folds of the sculpture appear with higher resolutions.}
	\label{fig:resolution}
	\vspace{-3mm}
\end{figure}

\begin{figure}[t]
	\centering
	\begin{overpic}[width=0.95\linewidth]{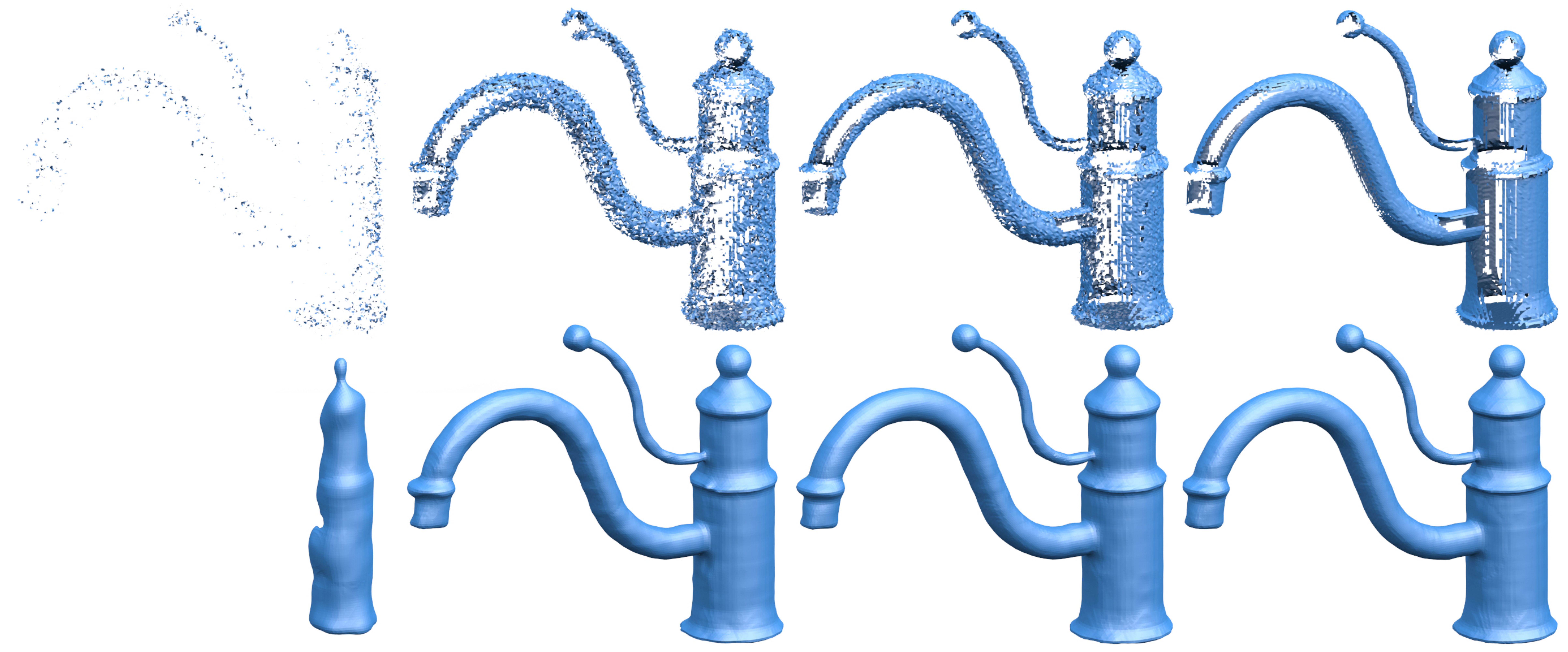}{
			\put(3,-2.7){\small (a) $\sigma=2e{-}3$}
			\put(28,-2.7){\small (b) $\sigma=4e{-}4$}
			\put(53,-2.7){\small (c) $\sigma=2e{-}4$}
			\put(80,-2.7){\small (d) $\sigma=0$}
	}\end{overpic}
	\vspace{0mm}
	\caption{Noisy input. By adding Gaussian noise of specified standard deviation $\sigma$ to depth maps, the partial scans (top row) exhibit increasing irregularity from right to left. The corresponding completed shapes (bottom row) are mostly stable against high frequency noise but cannot recover major structures once the partial scan has too few integrable signals in (a). $\sigma$ is defined relative to the object bounding box maximum side length.}
	\label{fig:noise}
	\vspace{-3mm}
\end{figure}
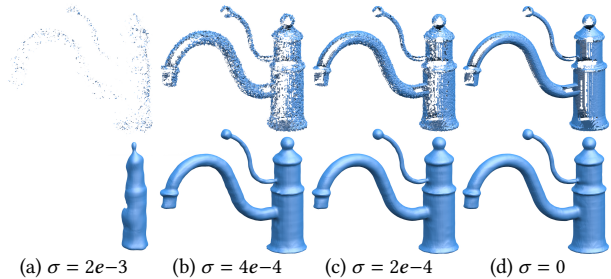

\paragraph{Resolution of volume} 
The implicit TSDF shape representation discretized on fixed resolution grids has difficulty in modeling very sharp features and intricate details. 
As shown in Fig.~\ref{fig:resolution}, with the increase of volume grid resolution, both the input partial scans and the result completed shapes have more subtle details.
However, higher resolution consumes larger memory space and longer computing time; 
our choice of $256^3$ balances between the computational power we have and the result quality.
In comparison, with sufficiently dense point cloud supervision, SPR can use locally supported RBF functions to capture fine details, and DGP using the powerful MLP mapping or Point2Mesh with the explicit triangle mesh representation can model sharp features more easily.
Our method is thus a useful addition to the existing reconstruction toolbox for recovering good overall shapes from incomplete input adaptively.

\paragraph{Input noise} 
We also test how the method works under noisy input. To this end, we add Gaussian noise $N(0,\sigma^2)$ to each pixel of the rendered depth maps independently, fuse the depth maps into the partial TSDF and complete with our method. 
As shown in Fig.~\ref{fig:noise}, with more noise added, the input partial scans first show high frequency irregularities over the surfaces but then lose major structures, due to the inconsistency of depth values across the views;
correspondingly, our method handles the high frequency noises without much degradation in result quality, but fails to recover main components once the major structures are lost in the input, which complies with observations made in the missing data ratio discussion above.
For the moderate noises, the robustness can be attributed to the SGD training dynamics which fit to the dominant global scale structures first and start to capture noise only at a later stage.
Indeed, more iterations on the noisy input lead to irregular reconstruction; see the supplemental document for a visual comparison.

\section{Conclusion}

We have presented a novel approach for single instance 3D shape completion, by using deep prior that automatically extracts the shape context in a hierarchical deep feature representation and completes the missing regions using similar patches in the context.
We interpret the deep prior from the neural tangent kernel perspective, to reveal that the deep prior effectiveness is a result of the deep CNN structure and the SGD training dynamics. 
Moreover, under guidance of the NTK interpretation, we design network structures and learning mechanisms that lead to more effective 3D completion by deep prior.
We evaluate the design principles with extensive ablation tests and visualizations.
Through comparisons with previous shape completion methods, our results are shown to be more aware of structures of various scales and complete larger missing regions, and complement the category supervised data driven methods by being more adaptive to shape instances.
It is a future work to explore how to combine our unsupervised single instance shape completion with the supervised data driven methods more efficiently.

\begin{acks}
	We thank Xin Tong and Yang Liu for discussions during the initial phase of the work, and the anonymous reviewers for suggestions.
\end{acks}

\appendix
\section{Appendix}
\label{sec:appendix}

\subsection{NTK derivation}
\label{sec:appendix:ntk_derivation}

First, we expand the gradient descent with respect to network parameters as
\begin{equation}
    \dv{\vb*{\theta}}{t} = -\nabla \ell(\vb*{\theta}) = \sum_{i=1}^{n}{\left(u_i - f(\vb*{\theta},\vb*{x}_i)\right)\pdv{f(\vb*{\theta},\vb*{x}_i)}{\vb*{\theta}}}.
\end{equation}
Then for any input $\vb*{x}$, we obtain the resulting dynamics of the network output as
\begin{align}
    \dv{f(\vb*{\theta},\vb*{x})}{t} = & \pdv{f(\vb*{\theta},\vb*{x})}{\vb*{\theta}} \dv{\vb*{\theta}}{t} \\
    = & \pdv{f(\vb*{\theta},\vb*{x})}{\vb*{\theta}}\sum_{i=1}^{n}{\left(u_i - f(\vb*{\theta},\vb*{x}_i)\right)\pdv{f(\vb*{\theta},\vb*{x}_i)}{\vb*{\theta}}} \\
    = & \sum_{i=1}^{n}{\left(u_i - f(\vb*{\theta},\vb*{x}_i)\right)\left<\pdv{f(\vb*{\theta},\vb*{x})}{\vb*{\theta}}, \pdv{f(\vb*{\theta},\vb*{x}_i)}{\vb*{\theta}}\right>}.
\end{align}

By assembling the labels and network outputs for the  training samples as $\vb*{u}=(u_i)_{i\in[n]}$ and $\vb*{y} = \left(f(\vb*{\theta}, \vb*{x}_i)\right)_{i\in[n]}$, the above formula can be further expressed as
\begin{equation}
    \dv{f(\vb*{\theta},\vb*{x})}{t} = \left(\textrm{ker}(\vb*{x},\vb*{x}_i)\right)_{i\in[n]}^T(\vb*{u} - \vb*{y}),
\end{equation}
where 
\begin{equation}
    \textrm{ker}(\vb*{x},\vb*{x}_i) = \left<\pdv{f(\vb*{\theta},\vb*{x})}{\vb*{\theta}}, \pdv{f(\vb*{\theta},\vb*{x}_i)}{\vb*{\theta}}\right>
\end{equation}
is the neural tangent kernel function.
We therefore have the kernel gradient update dynamics as
\begin{equation}
    \dv{\vb*{y}}{t} = \left(\dv{f(\vb*{\theta},\vb*{x}_i)}{t}\right)_{i\in[n]}
    = \vb{K}(\vb*{u} - \vb*{y}),
\end{equation}
as shown in Eq.~\ref{eq:kernel_gradient}, where $\vb{K}$ is the kernel Gram matrix.

\subsection{Network architecture details}
\label{appdx:network_architecture}

The network used for completion is composed of three hierarchies, where each hierarchy is an encoder/decoder structure without skip connection. 
One encoder block $e_{s,j}$ is made of 
$$- \text{Conv(2,2,0)} - \text{IN} - \text{LeakyReLU} - \text{Conv(3,1,1)} - \text{IN} - \text{LeakyReLU} - ,$$
and a decoder block $d_{s,j}$ is made of
\begin{align}
-&\text{Upsampling}-\text{IN}-\text{Conv(3,1,1)}-\text{IN}-\text{LeakyReLU}-\\
&\text{Conv(1,1,0)}-\text{IN}-\text{LeakyReLU}-,
\end{align}
where IN is instance normalization, and Conv(\textit{kernel size, stride, padding}) denotes a 3D convolution with specified kernel size, stride and padding size; the output channel sizes are given next.
We use three hierarchies each to deal with the input of a different resolution. 
The first hierarchy deals with the input of resolution $256^3$, and the feature size of encoder in this hierarchy is $[16,32,64,128,128]$ for $[e_{0,j}]_{j=1,\cdots,5}$, with the decoder feature size mirror-reflected; 
the second hierarchy deals with input that is down-sampled to $128^3$ by average pooling, with feature size $[16,32,64,128]$ for $[e_{1,j}]_{j=1,\cdots,4}$; 
the third hierarchy processes the input of $64^3$ with feature size $[16,32]$ for $[e_{2,j}]_{j=1,2}$. 
The input signal is a uniform noise sampled from $[0,0.1]$ with feature size 32.

\subsection{Notations used in the paper}
The notations used in the paper are summarized in Table~\ref{tab:notations}.
We use italic letters for scalars and functions, boldface lowercase letters for vectors, and boldface capital letters for matrices and tensors.

\begin{table}
	\vspace{-3mm}
	\begin{tabular}{r l}
		\\ \hline 
		Notation & Description \\ \hline
		$[n]$ & the sequence of integers from 1 to $n$ \\
		$\vb*{x}\in \mathcal{X}$ & an input sample $\vb*{x}$ in space $\mathcal{X}$ \\
		$f(\vb*{\theta},\cdot):\mathcal{X}{\rightarrow}\mathbb{R}$ & mapping  parameterized by $\vb*{\theta}$ \\
		$y=f(\vb*{\theta},\vb*{x})$ & output for input sample $\vb*{x}$ \\
		$u$ & output ground truth label \\
		$\vb{K}$ & kernel Gram matrix \\
		$\textrm{ker}(\vb*{x}_i, \vb*{x}_j)$ & NTK kernel function for a pair of inputs \\
		$\phi(\vb*{x})$ & NTK kernel feature map function \\
		$\Omega$ & domain of input TSDF \\
		$\mathcal{V}$ & regions known to be empty by scanning \\
		$\mathcal{M}$ & completion domain \\
		$\vb*{m}$ & mask of known regions of $\mathcal{M}$ \\
		$\vb{z}$ & input noise vector \\
		$\eta$ & the SDF clipping threshold\\
		$\vb{S}$ & the structure tensor of convolution\\
		$e_{s,j}$ & a block of the encoder at scale $s$ \\
		$d_{s,j}$ & a block of the decoder at scale $s$ \\
		\hline
	\end{tabular}
	\caption{Summary of notations used in the paper.}
	\label{tab:notations}
	\vspace{-8mm}
\end{table}

\bibliographystyle{ACM-Reference-Format}
\bibliography{completion}

\end{document}